%% file: paper-AAAI.tex
\def\year{2020}\relax
\documentclass[letterpaper]{article} 
\usepackage{aaai20}  
\usepackage{times}  
\usepackage{helvet} 
\usepackage{courier}  
\usepackage[hyphens]{url}  
\usepackage{graphicx} 
\usepackage{graphicx}  
\usepackage{amsthm}

\newtheorem{assumption}{A}
\input{preamble}

\title{Inferring Individual Level Causal Models from\\ Graph-based Relational Time Series}

\author{
Ryan Rossi,\textsuperscript{\rm 1}
Somdeb Sarkhel,\textsuperscript{\rm 1}
\Large \textbf{
Nesreen Ahmed\textsuperscript{\rm 2}
}%
\\
\textsuperscript{\rm 1}Adobe Research\\
\textsuperscript{\rm 2}Intel Labs
}
\begin{document}

\maketitle

\begin{abstract}
In this work, we formalize the problem of causal inference over graph-based relational time-series data where each node in the graph has one or more time-series associated to it. We propose causal inference models for this problem that leverage both the graph topology and time-series to accurately estimate local causal effects of nodes. Furthermore, the relational time-series causal inference models are able to estimate local effects for individual nodes by exploiting local node-centric temporal dependencies \emph{and} topological/structural dependencies. We show that simpler causal models that do not consider the graph topology are recovered as special cases of the proposed relational time-series causal inference model. We describe the conditions under which the resulting estimate can be used to estimate a causal effect, and describe how the Durbin-Wu-Hausman test of specification can be used to test for the consistency of the proposed estimator from data. Empirically, we demonstrate the effectiveness of the causal inference models on both synthetic data with known ground-truth and a large-scale observational relational time-series data set collected from Wikipedia.
\end{abstract}

\section{Introduction}\label{sec:intro}
Causal inference aims to estimate the effect of interventions and is fundamentally important for decision-making in many application domains~\cite{bakshy2014designing,bosch2012building}.
Most work on estimating causal effects (causal inference) assumes the data is independent and identically distributed (iid)~\cite{pearl2009causality}.
Under such assumptions, there has been a lot of work on using A/B testing, \ie, randomized experimentation for detecting such causal inferences.
However, most real-world data has non-iid structures, e.g., web pages have links, social media influencers have friends as well as followers. Recently, there has been work that extends such randomized experimentation techniques for the case of such structured data where instances may be connected~\cite{ugander2013graph}.
However, there are many practical settings where performing an experiment is impossible, impractical, expensive, or time-consuming.
For this reason, there has also been recent work on estimating causal effects from observational iid data~\cite{pearl1998graphs,cochran1973controlling,rosenbaum1984reducing,rubin2006matched}.
One particularly appealing class of methods for observational causal inference occurs when the observed data comes in the form of a time-series~\cite{gottman1978analysis,abadie2010synthetic}, where it is possible to infer the effect of an observed intervention on an individual unit when sufficient conditions are met. 

In order for accurate inference of effects in time series, it is is often necessary for practitioners to have access to a large number of pre-treatment time periods to ensure suitable accuracy of the employed models. 
%In many settings this requirement cannot be met, either due to a lack of observations or the applicability of only local stationarity assumptions for a time series model. 
In settings where the observed system consists of a \textit{network} of individuals this shortcoming is often addressed by employing the so-called templating assumption, where all nodes marginal and conditional structure are identical~\cite{arbour2016inferring}, and pooling observations for all units together.
However, in many settings observed effects can be heterogeneous with respect to the network structure, i.e., the causal effect of a treatment varies as a function of the position of the node in the network. 
As a motivating example, consider the case of Wikipedia page views after a major event such as an earthquake. 
The event is likely to have a dramatic effect on pages that are directly related to earthquakes, e.g. "earthquake" and "richter scale", but is unlikely to have an effect on distant pages in the Wikipedia page graph, e.g., the cooking page is unlikely to experience greater traffic. 
Unfortunately, neither the individual estimation method nor the templated model provide much utility in this setting. 
There is likely only a small amount of relevant observations on either side of the event time series making individual treatment under powered.
On the other hand, the number of nodes where unaffected implies that using the templated model will result in a diluted estimate of the causal effect. 

In this work, we provide a flexible alternative that allows for practitioners to strike a balance between individual level and fully templated models. 
Specifically, we assume that effects smoothly vary as a function of distance in the network to provide a mechanism that allows for weighted local pooling of observations which increases power. 
In this way, the model incorporates the fact that effects of a node are topologically dependent, \ie, if measurements are taken at different regions in the graph, then we would obtain different effect measurements.
Given an event of interest, e.g., the earthquake in our running example, this model can be employed within a interrupted time series design~\cite{gottman1978analysis} to infer causal effects. 

The remainder of this paper is organized as follows. 
We first review related work.
Next, we formally introduce the problem of relational time-series causal inference.
We then describe our approach for local estimation of node effects under this setting, and show how simpler models are recovered as special cases of the proposed causal inference model.
We examine the efficacy of our approach through a series of experiments on synthetically generated data.
Finally, we provide a demonstration on Wikipedia data assessing the effect of an earthquake on page views.

\section{Related Work} \label{sec:related-work}
Work related to ours can broadly be placed into the following categories: temporal relational learning, causal estimation in networks, and network diffusion processes. 

The closest work to our own is~\cite{marazopoulou2015learning}.
However, the approach in~\cite{marazopoulou2015learning} does not focus on estimating causal effects for relational time-series, and only focused on estimating the structure, rather than estimating the effects form a given causal structure. 
Previous work that leverages graph-based relational time-series data has focused primarily on classification
or regression, but do not provide a mechanism for causal inference in such relational time-series data.
In particular, existing work focuses on leveraging temporal dependencies to improve predictive performance~\cite{sharan2008temporal,rossi2012timespringer}.
However, this work focuses on classification and does not lend itself to producing causal quantities. 
Similarly, there has also been work on extending relational probability trees to the spatio-temporal domain for relational data that can vary in both space and time~\cite{mcgovern2008spatiotemporal}, but do not provide a mechanism for producing causal estimates. 
More recently, there has been work on relational time-series regression~\cite{rel-time-series-forecasting}, however, this work does not focus on causal inference.

There is a small but growing literature that concerns itself with modeling causal effects in relational data. 
\cite{arbour2016inferring} proposed relational covariate adjustment (RCA). RCA extends non-parametric adjustment to relational data, and allows the estimation of a wide range of functional dependencies.  
\cite{sherman2018identification} generalized non-parametric identification theory to latent variable causal chain graph models.
\cite{ogburn2018causal} proposed a parameterization for relational social network data, that corresponds to a particular family of graphical models known as
chain graphs. They demonstrated the potential of using chain graphs under certain conditions to analyze data with contagion and interference when DAG models are intractable.
\cite{ugander2013graph} proposed methods for A/B testing in the context of social networks where the treatment of individuals in the network spills over to neighboring individuals. Their proposed methods use graph clustering to partition the graph into clusters, and the estimate average treatment effects after adjustment under social interference.
A key difference between this paper and prior work is that here we explicitly take into account heterogeneity that is associated with the topology itself.
Prior work either corresponds to modeling only individual attributes, or assuming the global (templated) model. 

The final line of related work is work studying information diffusion processes, and contagion, e.g. \cite{leskovec2007dynamics,bakshy2011everyone,kempe2003maximizing,gomez2012inferring,rossi2012dpr-IM}.
Our work differs from this line in three important aspects. 
First and foremost, none of the above work deals with relational (graph-based) time-series data.
Second, the notion of heterogeneity based on topology can be seen as accounting for the effect of latent homophily in the networks--a distinct phenomenon from contagion. 
The second key distinction is in the estimand of interest. 
Finally, work in diffusion and contagion seeks to quantify aggregate measures of diffusion by considering individual effects, in contrast this work seeks to model individual effects as distinct estimands of interest.

\section{Problem Formulation} \label{sec:problem-formulation}
Throughout this work we employ the language of potential outcomes, where $X(0)$ indicates the counterfactual quantity, i.e., the value of $X$ that would have been observed had treatment been set to $0$. 
We define the \textit{individual effect} as the effect of an intervention on an individual's outcome, and the \textit{peer effect} as the effect of an intervention on an individual's immediate neighbors on the individual's outcome. We summarize the notation introduced in this paper in Table~\ref{table:notation}.

Assume $ G = \langle V, E \rangle$ is a graph with $n=|V|$ nodes and $m=|E|$ edges. Let $\mX$ be a $n \times t_{\max}$ matrix consisting of $n$ time-series of length $t_{\max}$.
\begin{equation}
    \mX^{\prime} = \mA \mX 
\end{equation}\noindent
Hence, $X^{\prime}_{it} = \sum_{j \in \Gamma_i} X_{jt}$ for any $t=1,\ldots,t_{\max}$ where $\Gamma_{i} = \{j \;|\; (i,j) \in E\}$ is the set of neighbors of node $i$.
Similarly, the relational time-series mean is:
\begin{equation} \label{eq:rel-temporal-mean}
    \bar{\mX} = \mD^{-1}(\mA\mX)
\end{equation}\noindent
where $\mD = \mathtt{diag}(\mA\ve) = \mathtt{diag}(|\Gamma_1|, \ldots, |\Gamma_n|)$ is the diagonal degree matrix $\ve$ is the vector of all ones.

We will assume that our estimand is the individual average treatment effect observed after making an intervention at time $t$, i.e. 
\begin{align*}
    \mathbb{E}\left(\sum_{k=t+1}^{t_{\max}} X_{ik}(1)\right) - 
    \mathbb{E}\left(\sum_{k=t+1}^{t_{\max}} X_{ik}(0)\right)
\end{align*}

We further assume that nodes whom obey some definition of closeness (proximity) have time series which evolve in a similar fashion, i.e. if two nodes are close in the network their temporal auto-dependence functions are also similar. 
In this setting, the problem at hand is to leverage nearby nodes to improve the statistical efficiency of the estimate for a particular node, i.e., 
we would like to learn a function as follows,
\begin{align*}
    \hat{X}_{ik} = g_i(X_{i,k-1}) + \sum_{j \in \Gamma_i} g_j(X_{i,k-1})
\end{align*}

In this work, we formalize the problem of estimating causal effects from \emph{relational time-series} data~\cite{rel-time-series-forecasting} as follows:
\begin{Definition}[\bf Relational Time-Series Causal Inference]
\label{prob:rel-time-series-causal-inference}
Let $ G = \langle V, E \rangle$ be a graph with $n=|V|$ nodes and $m=|E|$ edges; and let $\mX$ be a $n \times t_{\max}$ matrix of node time-series such that for each $i \in V$ there is an associated time-series $X_{i,1}, \dots X_{i,t_{\max}}$ of length $t_{\max}$.
We further assume that there is a known intervention that occurs at time $0 < t_{\textrm{int}} < t_{\max}$, and that the causal effect of the intervention is a smoothly varying process centered at some node $i \in V$.
The problem is to infer the individual and peer effects of node $i$.
\end{Definition}

Unless otherwise mentioned, we assume the following: 

\begin{assumption}\label{assump:boundeddeg}The maximum degree $\Delta(G) = \max\big\{ |\Gamma_1|, \ldots,|\Gamma_n|\big\}$ is bounded by some constant $c < \infty$\end{assumption}
\begin{assumption}For any three nodes, $i, j, k \in V$ if $i$ and $j$ share $k$ as a neighbor then the support of the conditional distributions $p(X_i | X_k)$ and $p(X_j | X_k)$ overlap, i.e. there is a shared support of the conditional distributions.\end{assumption}
\begin{assumption}No feedback cycles within time-steps\end{assumption}
\begin{assumption}There are no unobserved confounding variables.\end{assumption}

\begin{table}[t!]
\centering 
\caption{Summary of notation.}
\vspace{-2mm}
\centering 
\small
\setlength{\tabcolsep}{6pt} 
\def\arraystretch{1.18}
\label{table:notation}
\begin{tabularx}{1.0\linewidth}{@{}rX} 
\toprule
$G$ & a graph $G=(V,E)$ where $V$ is the set of nodes and $E$ is the edge set \\
$n$ & number of nodes in the graph, $n=|V|$  \\
$\mA$ & adjacency matrix of the graph $G$ \\
$\mX$ & node time-series matrix where each row is a node and columns represent time-series observations \\
$\vx_t$ & $n$-dimensional vector of time-series values for nodes at time $t$ \\
$X_{i,t}$ & time-series value of node $i$ at time $t$ \\
$\mD$ & diagonal degree matrix where $D_{ii}=|\Gamma_i|, \forall i\in V$ \\
$w$ & temporal lag (window size) \\
$d(i,j)$ & shortest path distance between nodes $i$ and $j$ in $G$ \\
$\gamma$ & a decay factor that determines the weight given to nodes (and their time-series) further away in the graph \\
$\beta$ & estimated regression coefficient \\
$\Gamma_i$ & set of neighbors of node $i$ \\
$t_{\max}$ & the last observation in the time-series \\
\bottomrule
\end{tabularx}
\end{table}

\section{Framework}
In this section, we describe our proposed approach, then we describe how current alternative (individual and global/templated models) can be viewed as special cases of the framework. 
Without loss of generality, we will assume linear models throughout this section. 
Consider the following estimation problem for a single node $j$:
\begin{align}
\label{opt:individual}
    \min_\beta \sum_i^n \sum_{t \in 2\dots t_{\textrm{max}}} (X_{i, t} - \beta X_{i, t-w})^2 \gamma^{d(i, j)}
\end{align}
where $w$ is the temporal lag (window size) and $d(i, j)$ is defined as the shortest path distance between nodes $i$ and $j$ in the graph.
In this problem $\gamma$ provides the ability to control the extent to which information from other nodes in the network contribute to the estimation of $\beta$.
At one extreme, as $\gamma$ approaches zero, the estimate will recover an i.i.d. estimate. 
At the other, as $\gamma$ approaches 1, the estimate will pool all instances and a global model is recovered.

A similar definition can be made in the case of measuring peer-effects. 
Here, we will assume that that the estimand of interest is the global average treatment effect, i.e., the counterfactual is that \textit{both} the individual and their peers are treated. 
We can now define an analogous quantity for the global treatment effect where we have defined $\beta^I$ as the coefficient for individual effects and $\beta^P$ as the coefficient for peer values. 
\begin{align}
\label{opt:individual}
    \min_\beta \sum_{i \in \Gamma_j} \sum_{t \in 2\dots t_{\textrm{max}}} (X_{i, t} - \beta X_{i, t-w})^2 
\end{align}\noindent
where $\Gamma_j = \{i \in V | (i,j) \in E\}$.

\begin{align}
\label{opt:peer}
    \!\!
    \min_{\beta^I, \beta^P} \sum_{i}^n \!\sum_{t \in 2\dots t_{\textrm{max}}} 
    \!\!\left(X_{i, t} \!\!- \!\left(\beta^I X_{i, t\!-\!w} \!\! +\! \beta^P \big[\mD^{\!-\!1}\!\!\mA_{}\mX\big]_{i,t\!-\!w}\! \right)\!\right)^2 \! \gamma^{d(i, j)}
\end{align}\noindent
where $\big[\mD^{-1}\mA_{}\mX\big]_{i,t\!-\!w}$ is the average peer value at time $t-w$.
Both optimization problems in Eq.~\ref{opt:individual} and Eq.~\ref{opt:peer} are easily implemented with off-the-shelf software packages by giving each sample a weight defined by $\gamma^{d(\cdot, \cdot)}$.

Existing work modeling causal effects in temporal non-relational (iid) data are recovered as special cases of the proposed approach.
We now address each in turn. 

\medskip\noindent\textbf{Individual Model}. 
This model estimates causal inference using only the time-series of the specific node.
This approach reduces the problem to standard multivariate time-series modeling~\cite{akaike2012practice}.
The advantage of this approach is that it allows the practitioner to use long-studied methodology and is essentially assumption free with respect to the form of relational dependence. 
Within the framework described above, this model corresponds to setting $\gamma = 0$.
The corresponding effect models are given by
\begin{align}
\label{opt:individual-iid-model}
    \min_\beta \sum_{t \in 2\dots t_{\textrm{max}}} (X_{j, t} - \beta X_{j, t-w})^2 
\end{align}\noindent
\noindent\textbf{Individual (IID) Peer Model}:
Similarly, if we set $\gamma=1$, then the local peer model (Eq.~\ref{opt:peer}) reduces to the individual (iid) peer model:
\begin{align}
\label{opt:peer-individual-iid-model}
    \!\!
    \min_{\beta^I, \beta^P} \sum_{t \in 2\dots t_{\textrm{max}}} 
    \!\!\left(X_{j, t} \!\!- \!\left(\beta^I X_{j, t\!-\!w} \!\! +\! \beta^P \big[\mD^{\!-\!1}\!\!\mA_{}\mX\big]_{j,t\!-\!w}\! \right)\!\right)^{\!2}
\end{align}\noindent
The cost of this approach, however, is that the effective sample size of the data is limited by the number of observations. 
This trade-off is often untenable for practitioners who work with moderately sized time series and/or are seeking to measure phenomenon which are characterized with small effect sizes.

\medskip\noindent\textbf{Global Model}
On the other side of the spectrum is assuming a global, i.e., templated model.
In this approach all nodes are assumed to have the same marginal and conditional distributions, all observations are pooled and treated as if they were generated from a single node and it's neighbors (in the case of peer effect measurements), which corresponds to setting $\gamma = 1$ in our proposed framework.
The corresponding individual and peer effect models are given by
\begin{align} 
\label{opt:individual-global-model}
    \min_\beta \sum_i^n \sum_{t \in 2\dots t_{\textrm{max}}} (X_{i, t} - \beta X_{i, t-w})^{2} 
\end{align}
\noindent\textbf{Global Peer Model}:
Similarly, if we set $\gamma=0$, then the local peer model (Eq.~\ref{opt:peer}) reduces to the global peer model:
\begin{align}
\label{opt:peer-global-model}
    \!\!\!\!\!
    \min_{\beta^I, \beta^P} \sum_{i}^n \!\sum_{t \in 2\dots t_{\textrm{max}}} 
    \!\!\!\left(\!X_{i, t} \!\!- \!\!\left(\beta^I \! X_{i, t\!-\!w} \!\! +\! \beta^P\! \big[\mD^{\!-\!1}\!\!\mA_{}\mX\big]_{i,t\!-\!w}\! \right)\!\right)^{\!2}
\end{align}\noindent
The benefit of this approach is that the number of observations available for a single model can be used to infer parameters. 
Unfortunately, the templating approach comes at a significant cost in terms of the necessary assumptions, which are opaque and difficult to reason over for practitioners. 
It is straightforward to see that the global and individual (iid) models are special cases of the proposed \emph{local causal inference models}.

\subsection{Test of Specification}
The consistency of the proposed local modeling assumption hinges on the level of correlation between the target node series and the series of the rest of the network and the value of $\gamma$.
If $\gamma$ is set to be too high with respect to the level of correlation between the series the resulting estimate can be very biased. 

Since the individual model is consistent, the Durbin-Wu-Hausman test~\cite{durbin1954errors} can be used to test correct specification of the proposed model as follows:
\begin{align}
    &H(\beta_{local}) = \\
    &(\beta_{local} \!- \beta_{ind})^T(\Var(\beta_{ind}) \!-\! \Var(\beta_{local}))^{-1}(\beta_{local} \!- \beta_{ind}) \nonumber
\end{align}
A test of significance can be performed by using the Chi-Square distribution with the degrees of freedom given by the rank of $(\Var(\beta_{ind}) - \Var(\beta_{local}))$.
By having a test of specification practitioners can make a principled decision on choosing between the local and individual level data without relying on opaque assumptions.

\begin{figure*}[h!]
\centering
\subfigure{\includegraphics[width=0.33\linewidth]{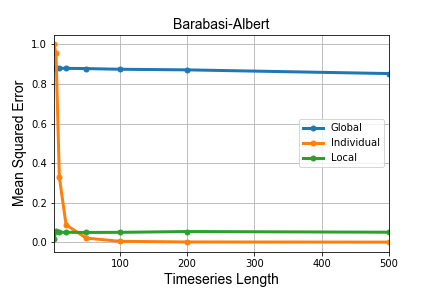}}
\subfigure{\includegraphics[width=0.33\linewidth]{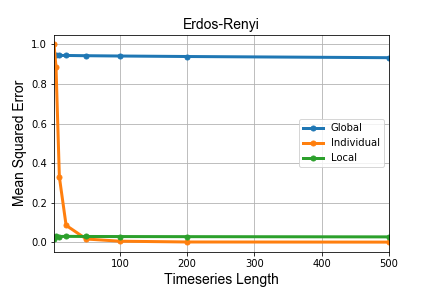}}
\subfigure{\includegraphics[width=0.33\linewidth]{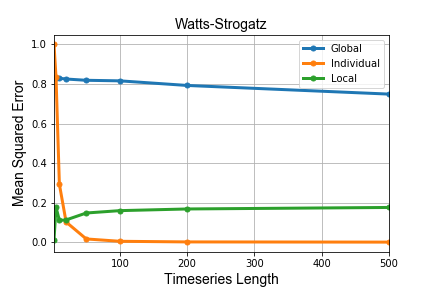}}

\caption{
Mean squared error~($(\beta - \hat{\beta})^2$) under different  network models for inferring an individual effect given heterogeneously affected neighbors.
The local model provides the lowest error estimates in small sample sizes, and significantly less biased estimates as the sample size is increased. 
See text for more discussion.
}
\label{fig:synthind}
\end{figure*}
\begin{figure*}[h!]
\centering
\subfigure{\includegraphics[width=0.33\linewidth]{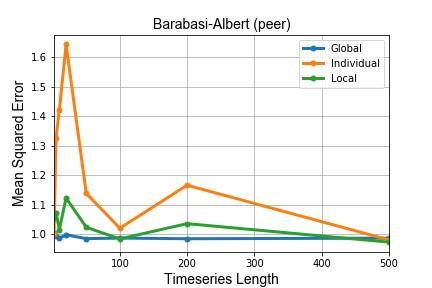}}
\subfigure{\includegraphics[width=0.33\linewidth]{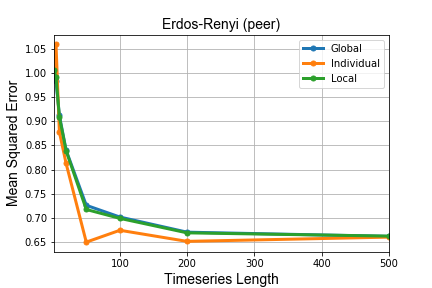}}
\subfigure{\includegraphics[width=0.33\linewidth]{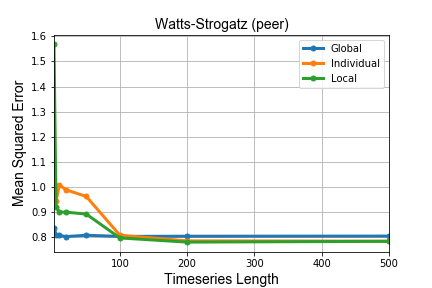}}

\caption{Mean squared error~($(\beta - \hat{\beta})^2$) under different network models for inferring a peer effect given heterogeneously affected neighbors.
While all models converge to the same level of error with larger number of time steps, the local model provides an estimate with lower error than the individual only model at smaller lengths. 
This, in concert with the results in figure \ref{fig:synthind}, indicate that the local model provides a robust and flexible approach for inference. See text for a full discussion.}
\label{fig:synthpeer}
\end{figure*}

\subsection{Inferring Causal Effects}
Thus far we have focused on modeling the relational time series, but have yet to address how a causal quantity can be obtained.
Given the proposed setting of a relational time series and an observed intervention, a interrupted time series design~\cite{hahn2001identification}, can be employed to estimate the effect under the following assumption:
\begin{assumption}
\label{assump:constant}
For all nodes with $\gamma > 0$, the effect of treatment is constant across units.
\end{assumption}
We note that while assumption \ref{assump:constant} is strong, the condition can be checked by applying the Hausmann test as described previously. 
The interrupted time series design in the individual model corresponds to,
\begin{align*}
    X_{i, t} \sim \beta^I X_{i, t-w} + \beta^C c X_{i, t -w} + \beta^{C'} c + \epsilon
\end{align*}
where $c$ is an indicator function of treatment and $\beta^C$, $\beta^{C'}$ is the post treatment offset and slope, respectively. 
Modeling of both peer and individual effects simultaneously can be achieved by considering an interaction between treatment status and the first order time series differences.

\section{Experiments} \label{sec:exp}
In this section, we first validate the causal models using synthetic data with known ground-truth and then investigate using the causal inference models for estimating local node effects from a real large-scale relational time-series data set.

\subsection{Synthetic Graph Experiments} \label{sec:exp-syn}
We first evaluate the efficacy of the proposed approach on synthetic data. 
Throughout we consider networks generated using the following graph models:
\begin{enumerate}
\item Erdos-Renyi~(Random) with probability of edge existence varied between $[0.1, 0.3]$.
\item Watts-Strogatz~(Small-World) with neighborhood size of 5 and rewiring probability set within
to $[0.15, 0.3]$
\item Barabasi-Albert~(Scale Free) with power of preferential attachment varied from 1 to 5. 
\end{enumerate}
We consider two ground-truth scenarios. 
In the first, we seek to measure the individual effect, with each observation at time $i$ generated as $$x_i \sim \beta x_{i - 1} + \epsilon$$
where 
$$\epsilon \sim \mathcal{N}(0, 1)$$
Each node's $\beta$ is given as a draw from a multivariate normal with mean 1, and covariance given by transition probability from nodes $i$ to $j$ after 2 steps. 
Finally, we consider the case of heterogeneous effects with the addition of peer effects. 
In the second scenario we consider the case of peer effects, with each observation at time $i$ generated as
$$x_i \sim \beta^I x_{i - 1} + \beta^P \mD^{-1}\mA x_{i-1} + \epsilon$$
where
$$\epsilon \sim \mathcal{N}(0, 1)$$
For all methods we ran 5000 trials and report the root mean squared error of the estimated causal effect from ground truth, i.e. $\|\beta - \hat{\beta}\|^2$, where $\hat{\beta}$ is the inferred coefficient. 
We compare against a model that models each node's series independently~(\textbf{ind}) and a templated model~(\textbf{relational}).
For the local model we consider $\gamma = 0.05$, which we have found works well across a variety of settings, using cross-validated estimates. 

The results can be seen in Figures~\ref{fig:synthind} and~\ref{fig:synthpeer}. 
We see that across graph topologies, the local estimate provides the lowest error when a small number of observations ($<$50) is available. 
When the time series length increases, the individual model begins to perform better, which is expected given the simplicity of the model and the known consistency of the individual model for large samples. In the case of peer effects we see that all estimates eventually converge, but the individual level model has much poorer performance in moderate sample sizes.
This demonstrates the advantages of local pooling: the increased power associated with incorporating local observations decreases the bias substantially in small sample sizes, and by considering only nearby observations we avoid the substantial bias that can result from global pooling. 

\subsection{Observational Network Data Experiments} \label{sec:exp-real-network}
We also investigate the causal inference models using real-world network data extracted from Wikipedia consisting of 4,143,840  Wikipedia pages (nodes) with 72,718,664 hyperlinks (edges) between those pages.
In addition to the large Wikipedia hyperlink graph described above, we also obtained a time-series for each page (node) representing the hourly page views, \ie, the number of times a page was viewed in a given hour.
In other words, each node in the graph has an associated (relational) time-series of hourly page views.
There are a total of 48 hours of page view time-series data starting from March 6, 2009 and moving forward in time.
In that time period, the average page views is 1.42 whereas the maximum page views is 353,799 for any page at any time.
The time-series of page views for each node can be interpreted as a measure of external interest in Wikipedia pages.

\begin{figure}[b!]
\vspace{-3mm}
\centering
\includegraphics[width=0.8\linewidth]{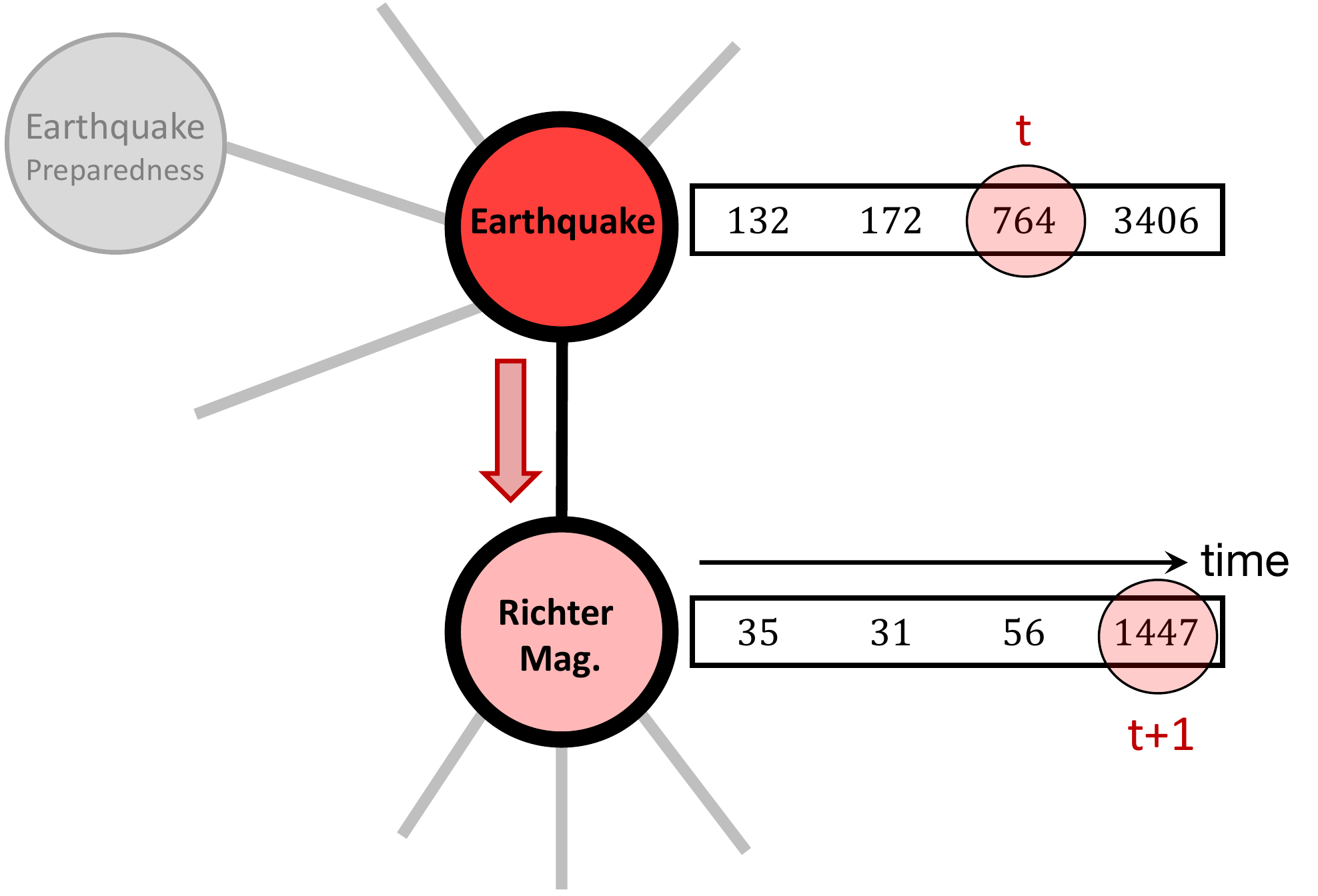}
\caption{Example with known causal effect between the Earthquake and Richter Magnitude pages.}
\label{fig:earthquake-example-known-causal-effect}
\end{figure}

\begin{figure*}[h!]
\centering
\subfigure{\includegraphics[width=0.245\linewidth]{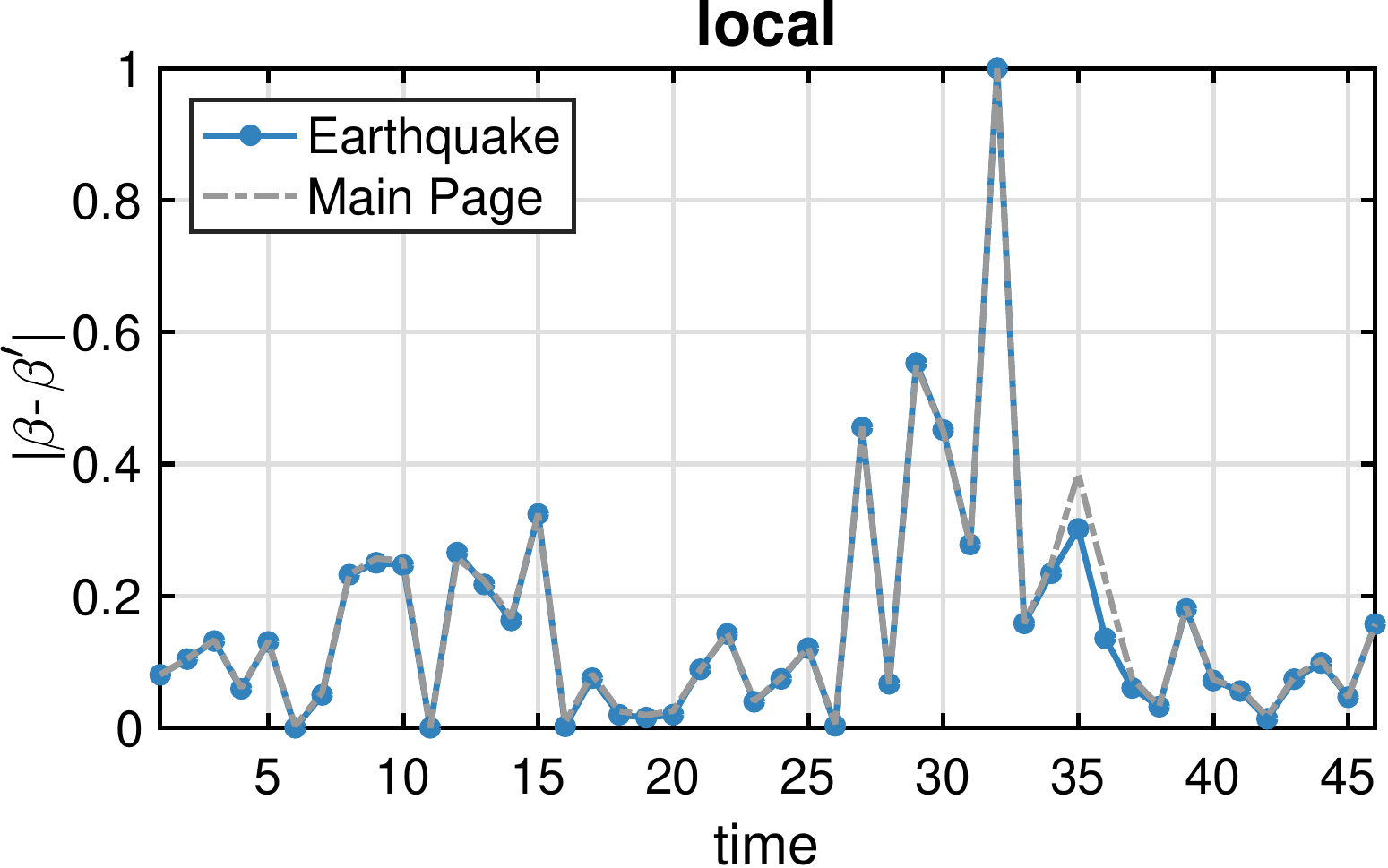}}\hfill
\subfigure{\includegraphics[width=0.245\linewidth]{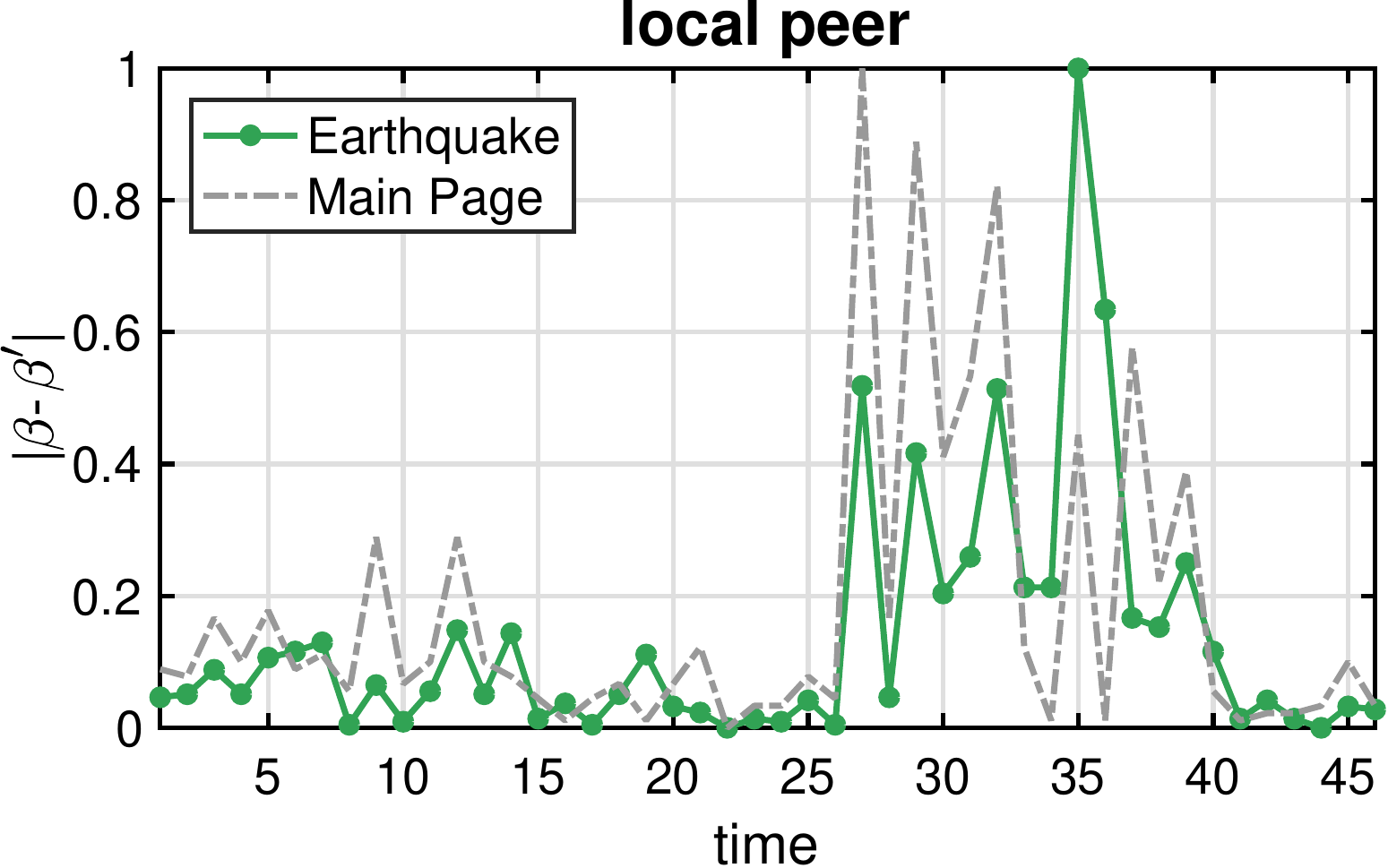}}\hfill
\subfigure{\includegraphics[width=0.245\linewidth]{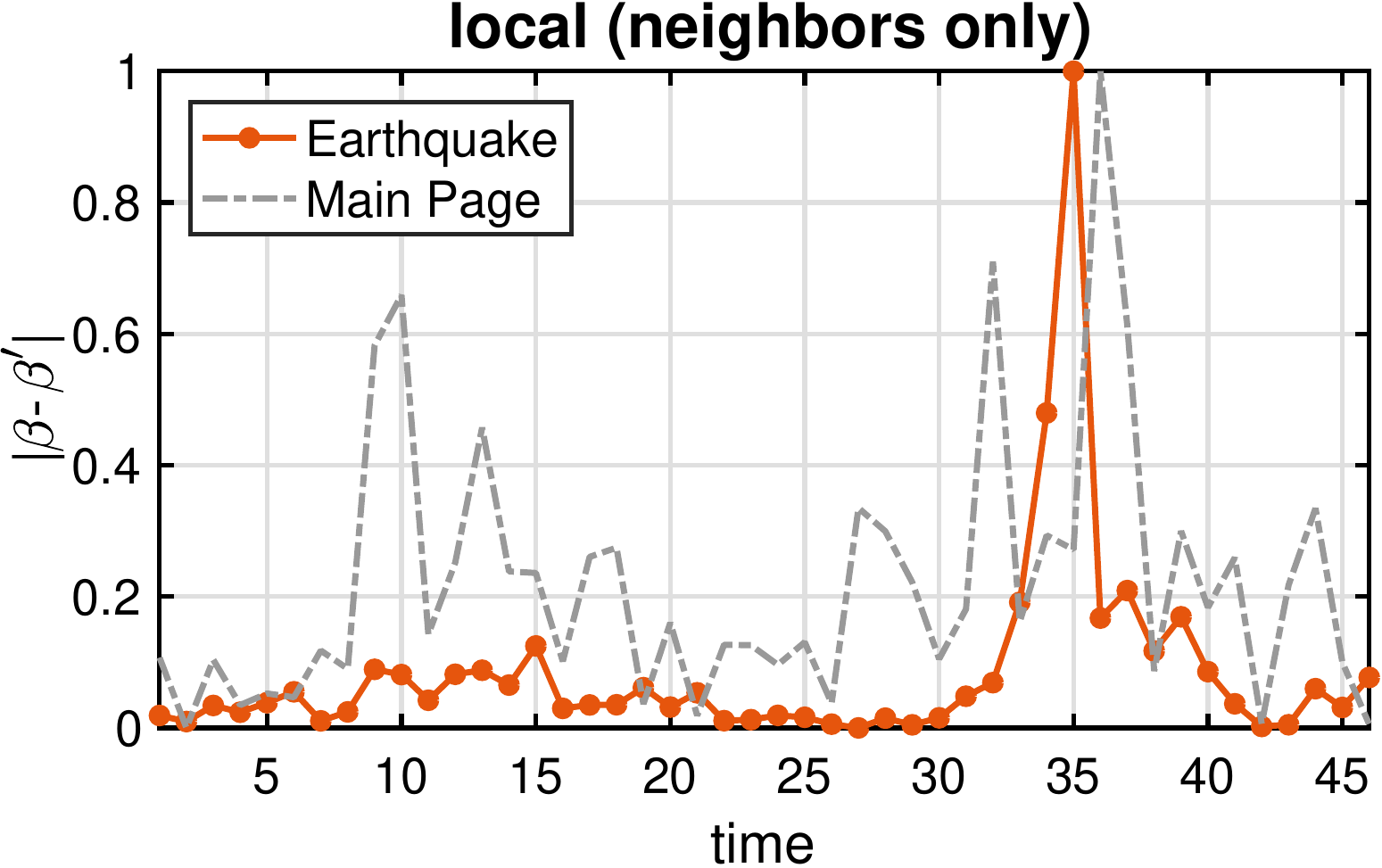}}\hfill
\subfigure{\includegraphics[width=0.245\linewidth]{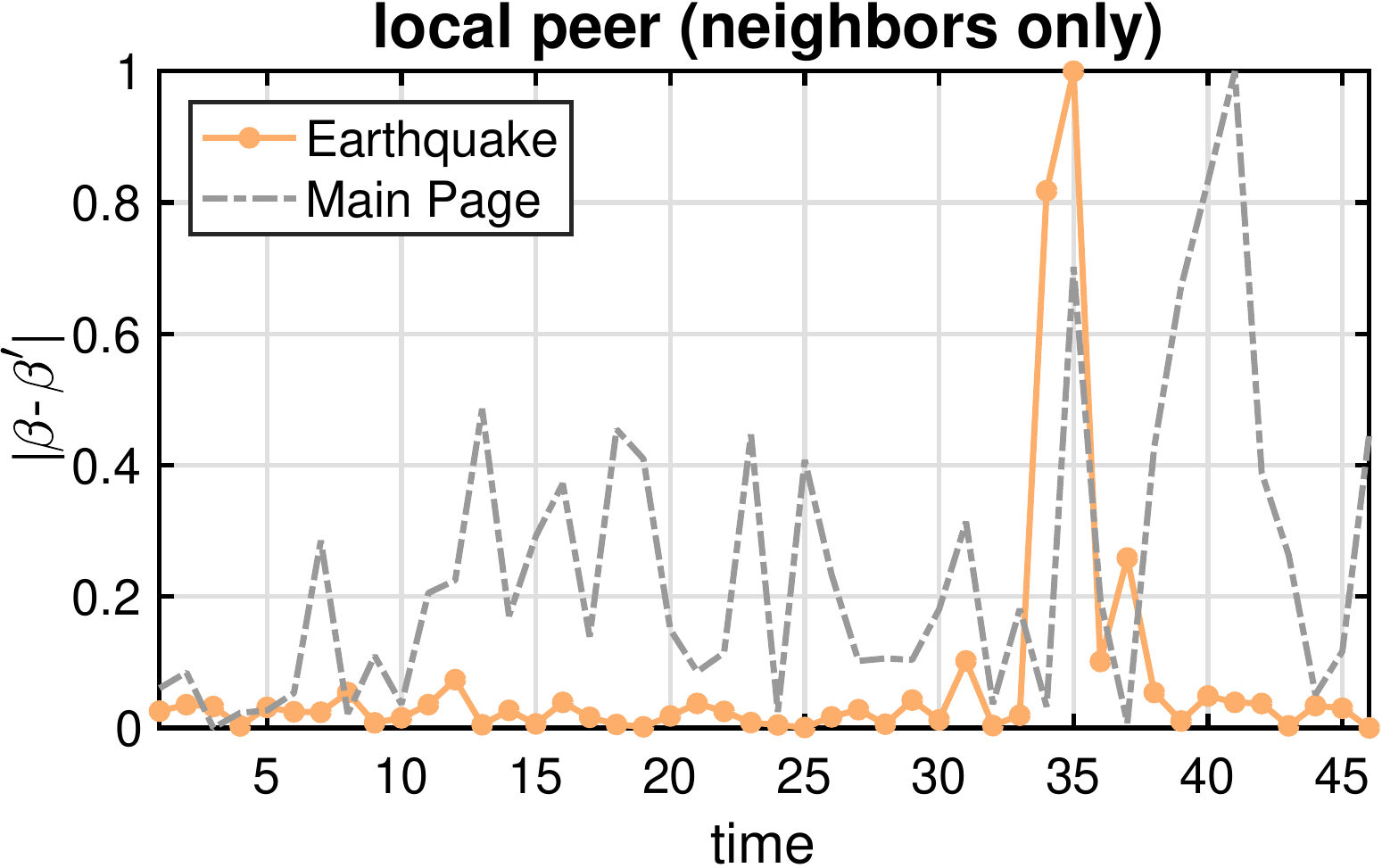}}\hfill

\subfigure{\includegraphics[width=0.245\linewidth]{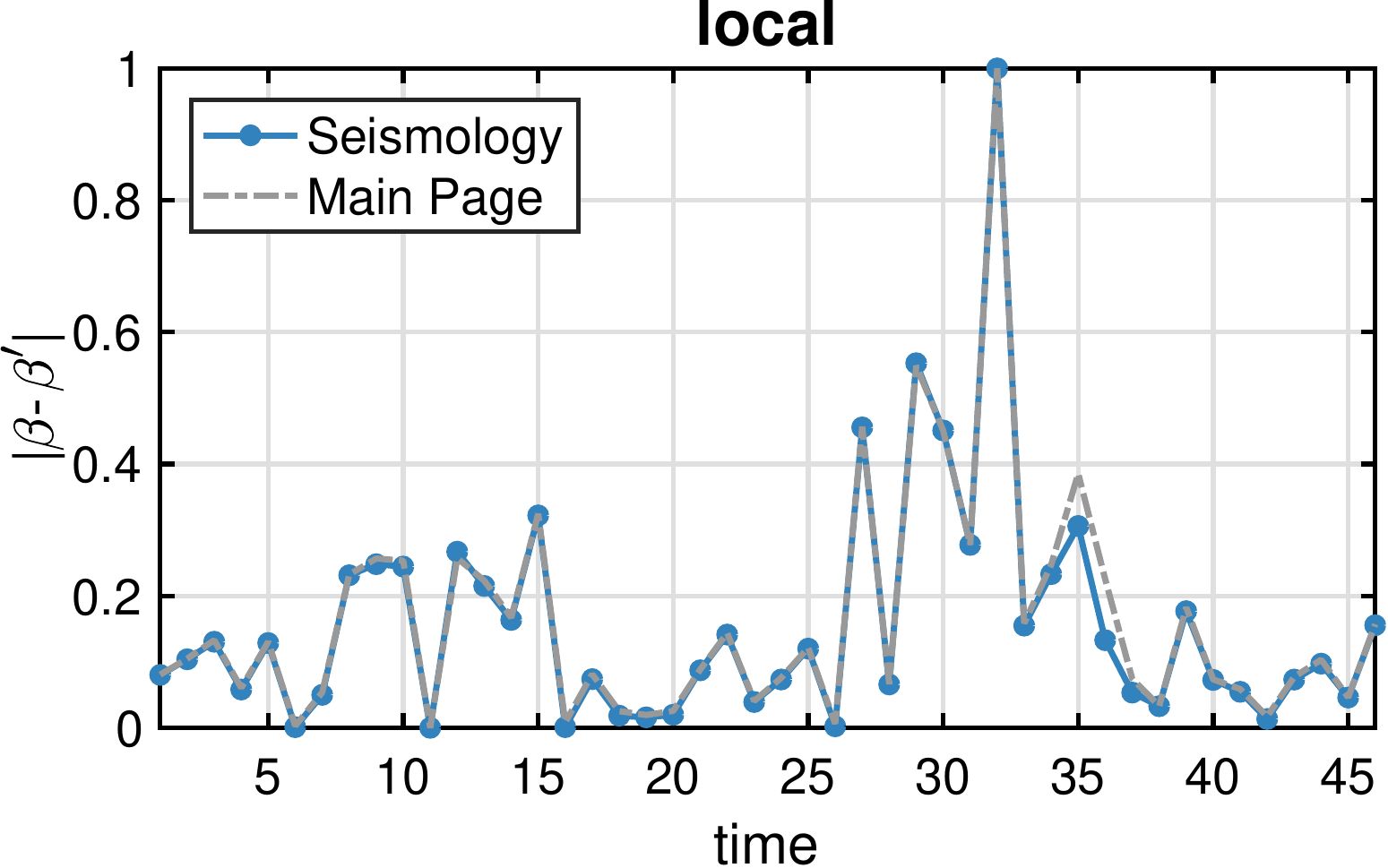}}\hfill
\subfigure{\includegraphics[width=0.245\linewidth]{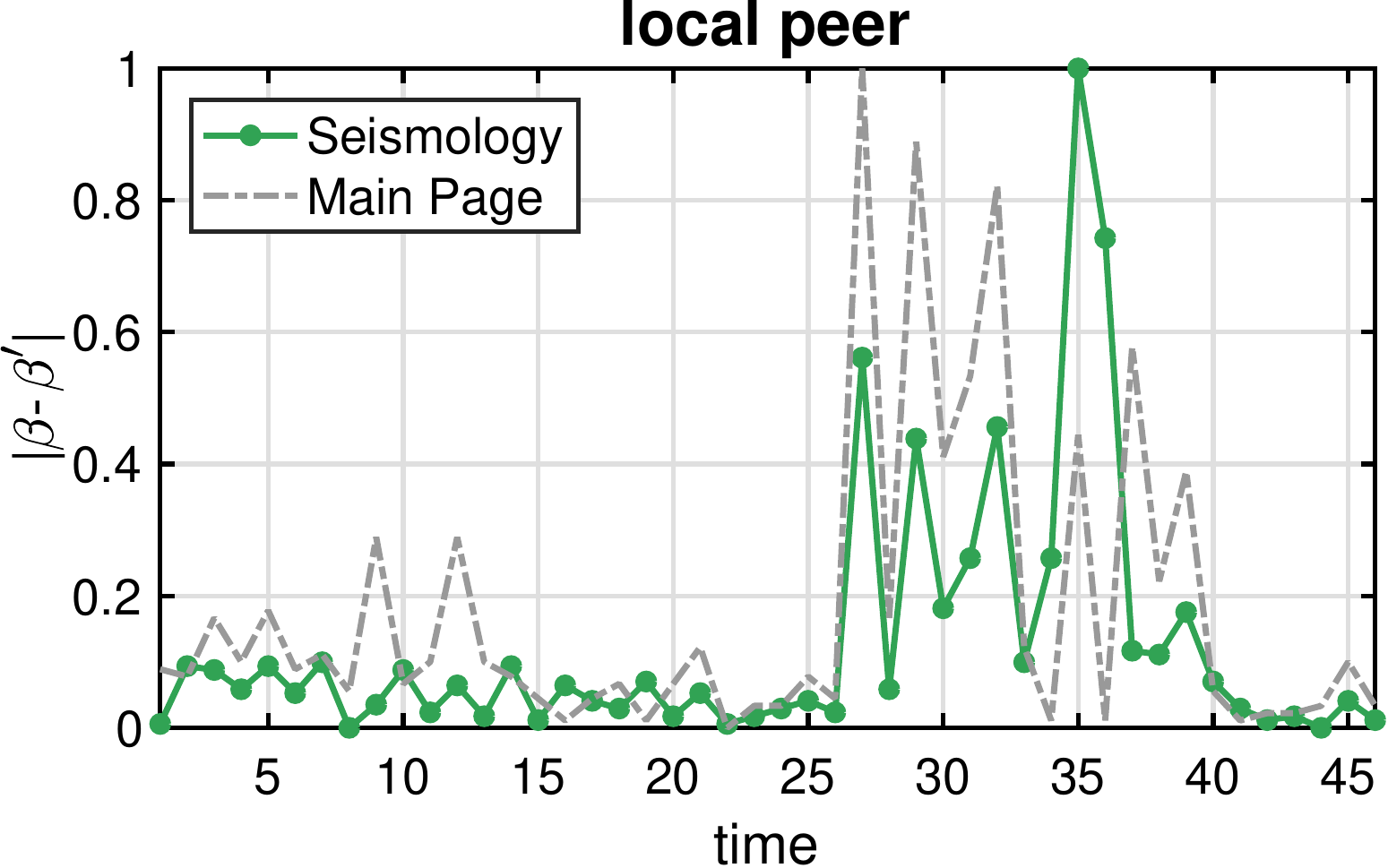}}\hfill
\subfigure{\includegraphics[width=0.245\linewidth]{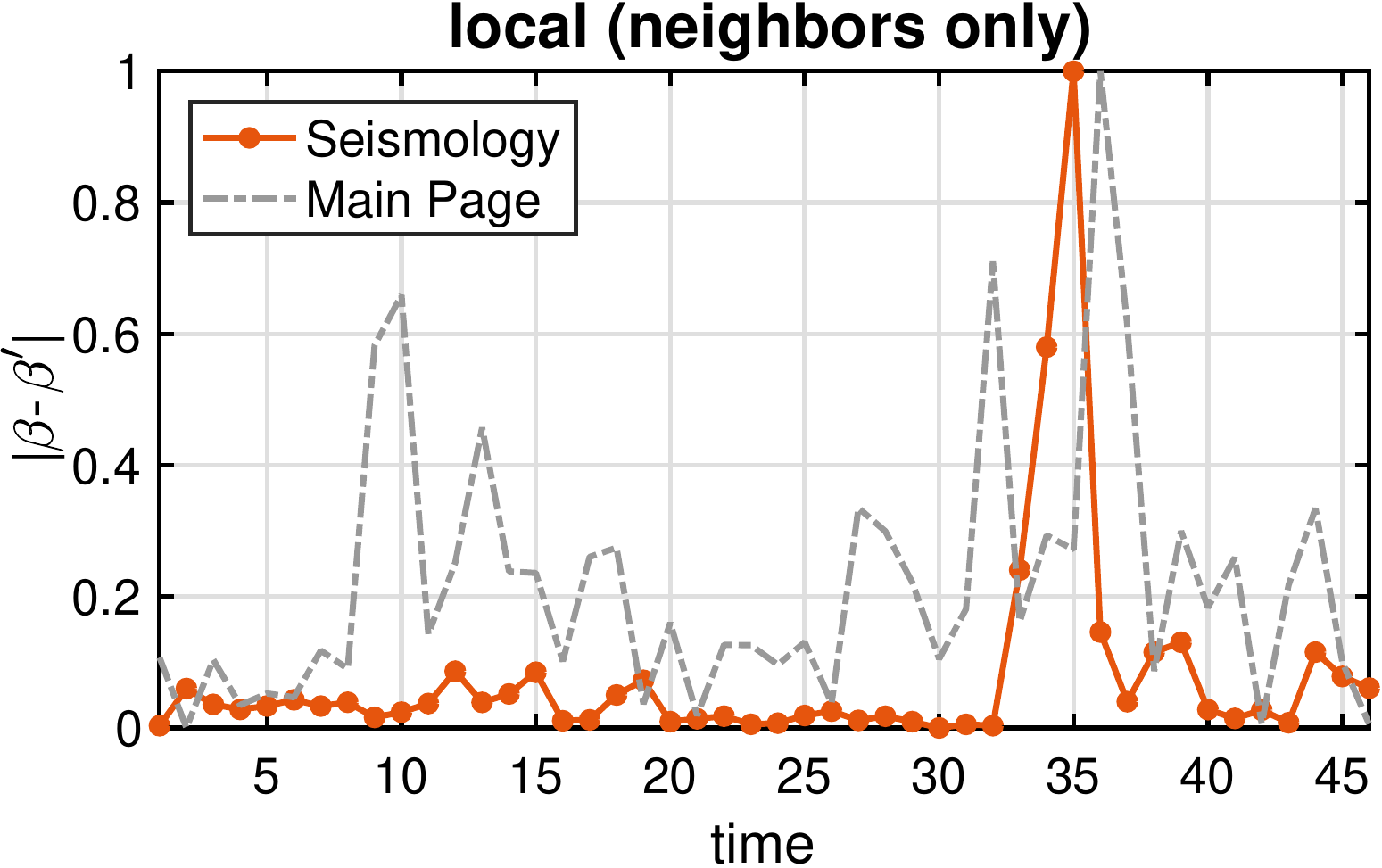}}\hfill
\subfigure{\includegraphics[width=0.245\linewidth]{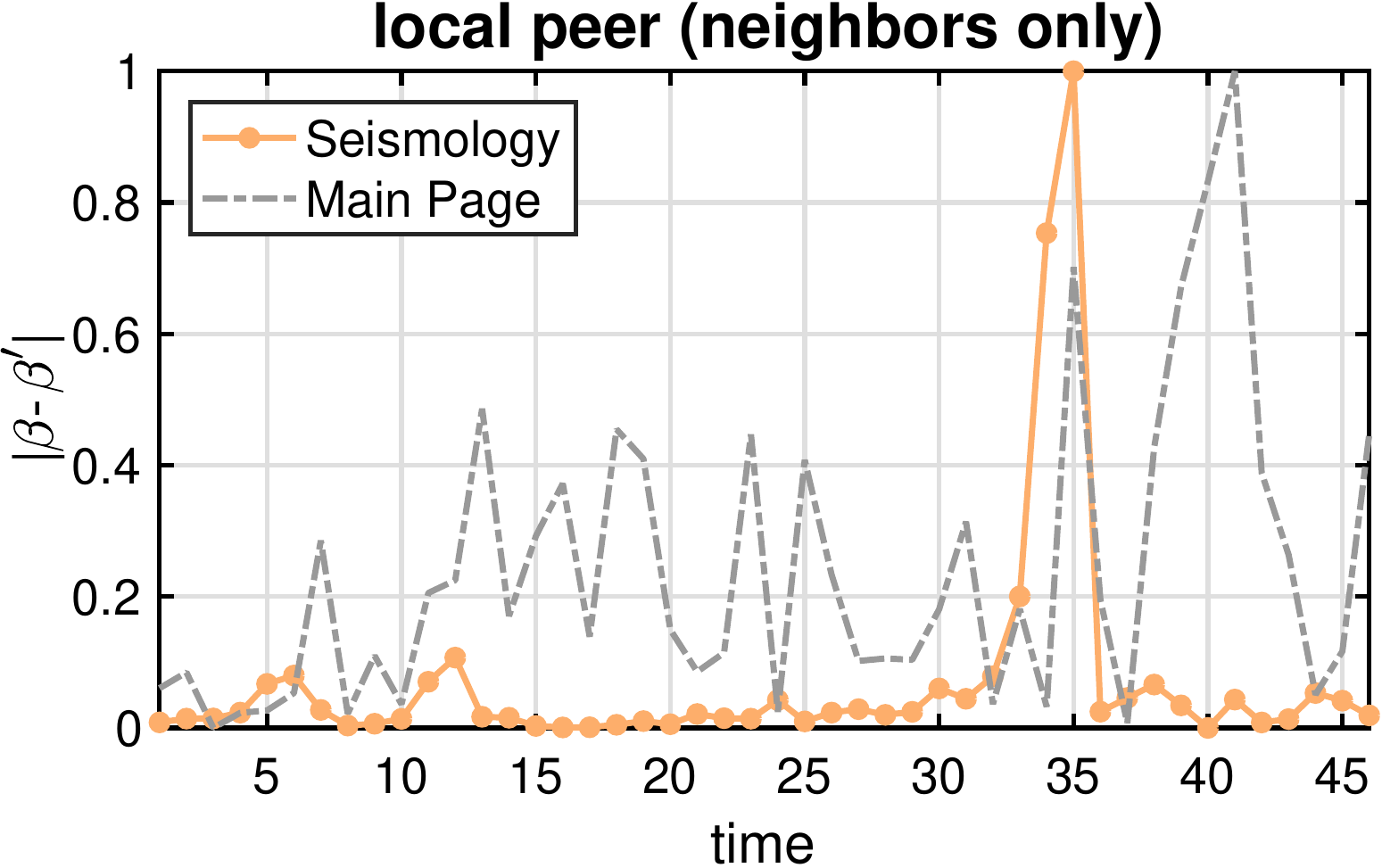}}\hfill

\subfigure{\includegraphics[width=0.245\linewidth]{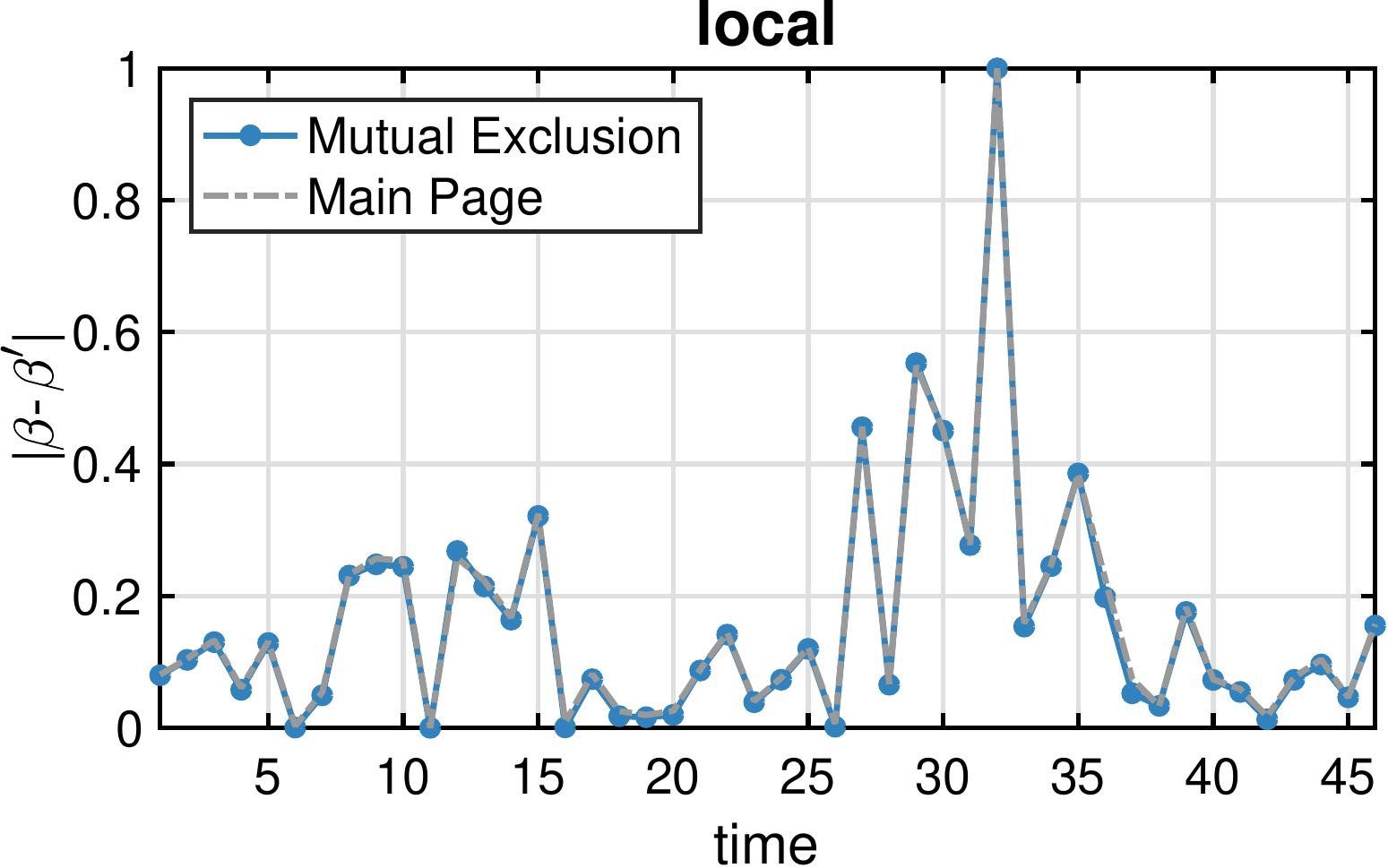}}\hfill
\subfigure{\includegraphics[width=0.245\linewidth]{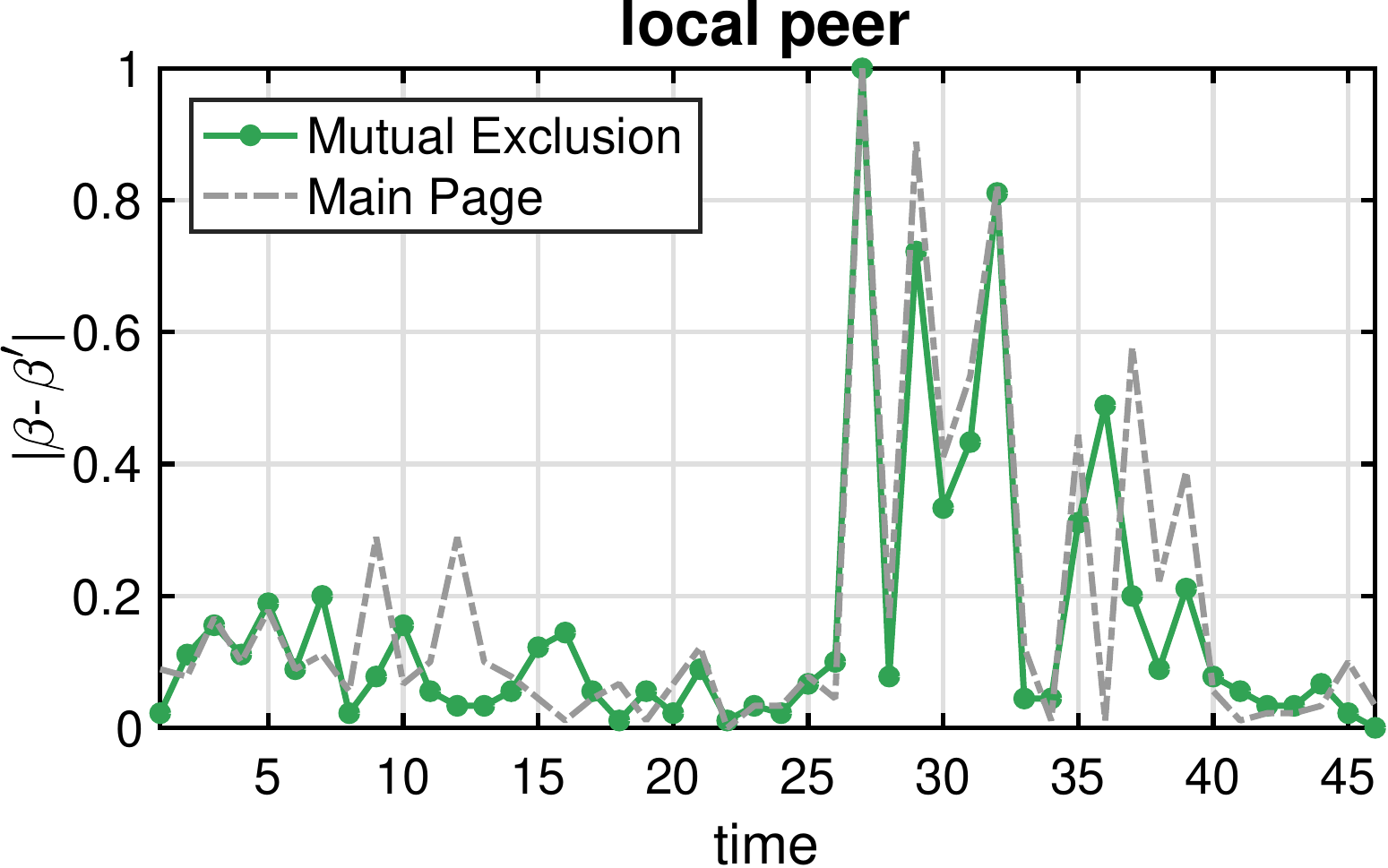}}\hfill
\subfigure{\includegraphics[width=0.245\linewidth]{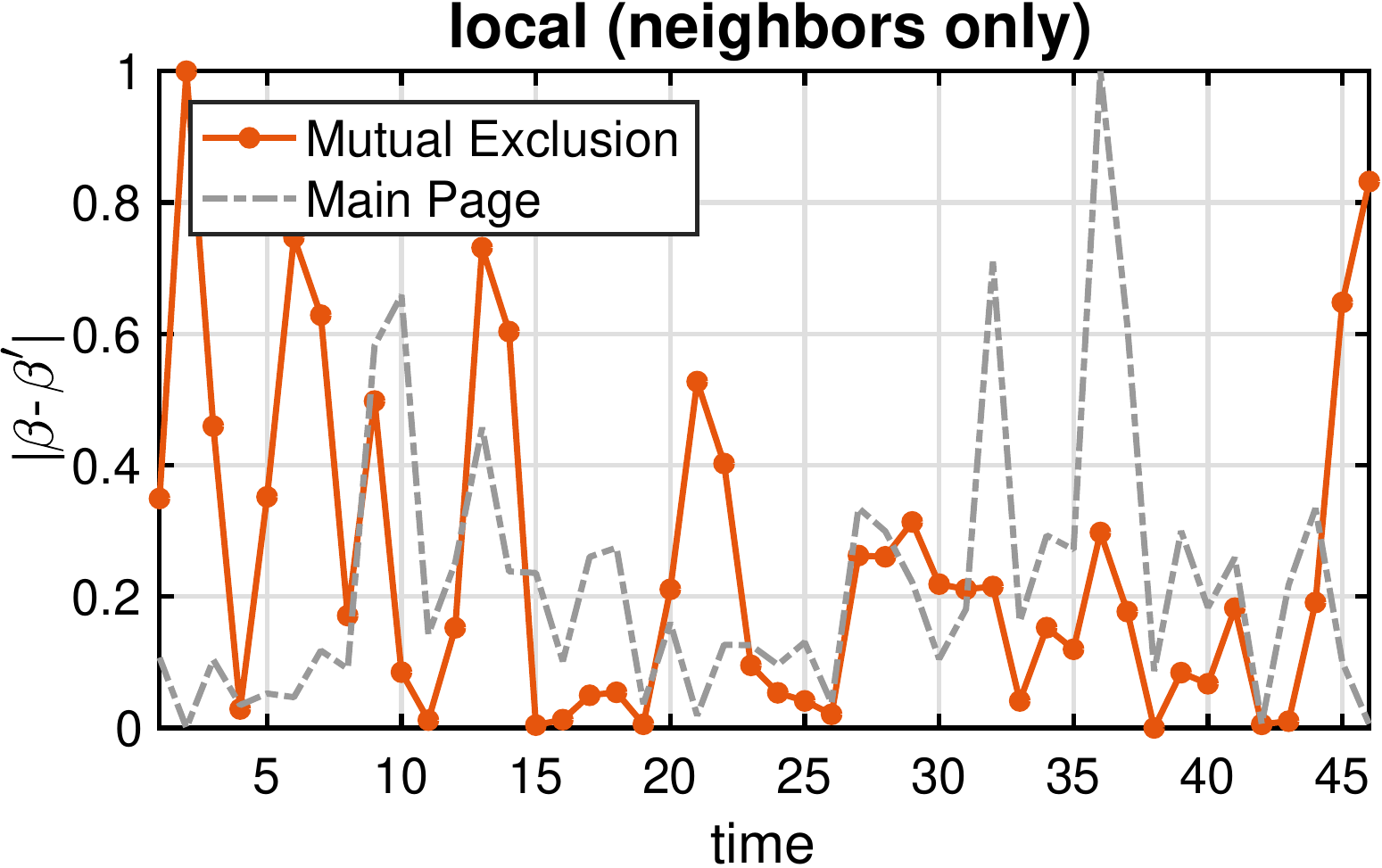}}\hfill
\subfigure{\includegraphics[width=0.245\linewidth]{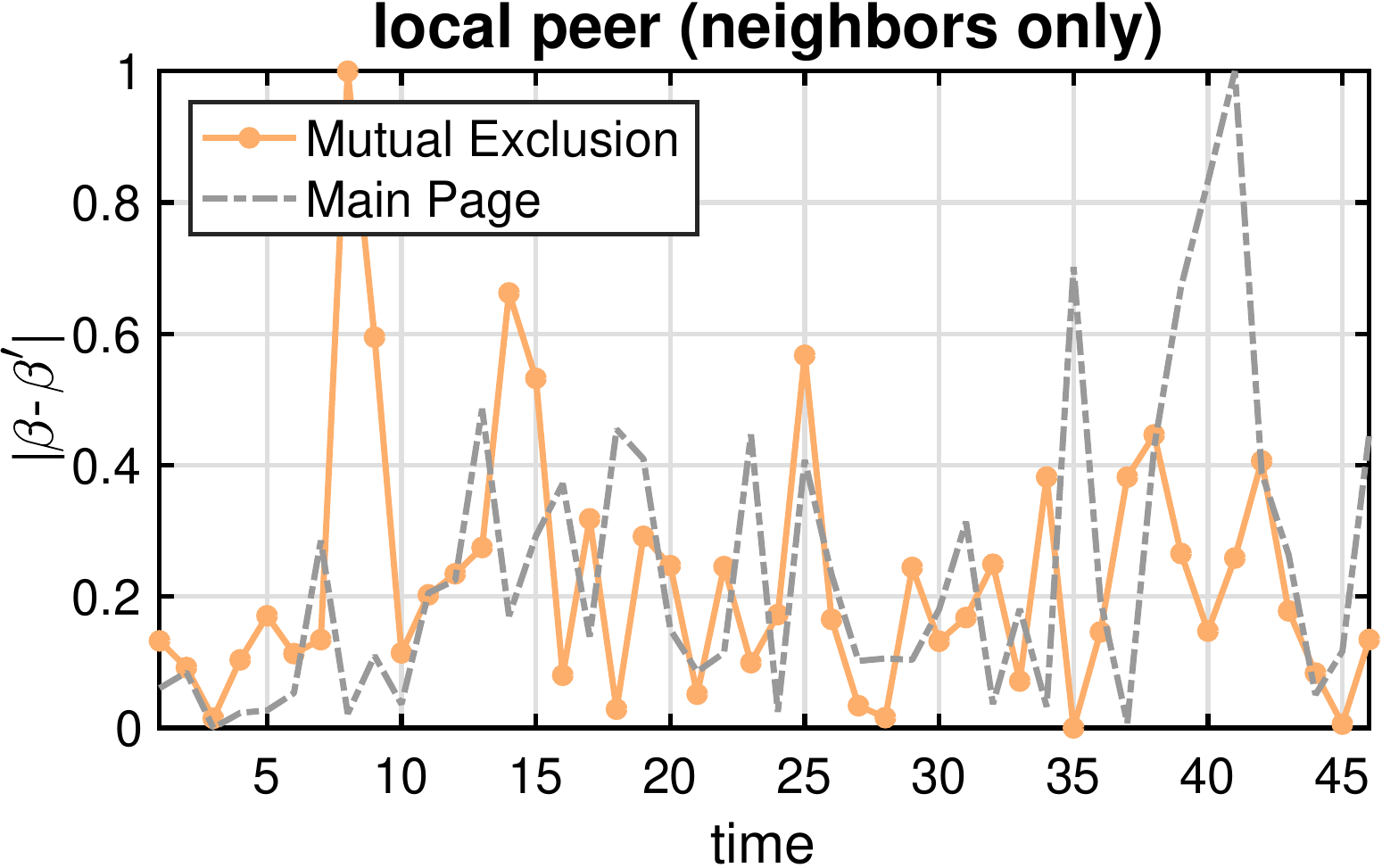}}\hfill

\subfigure{\includegraphics[width=0.245\linewidth]{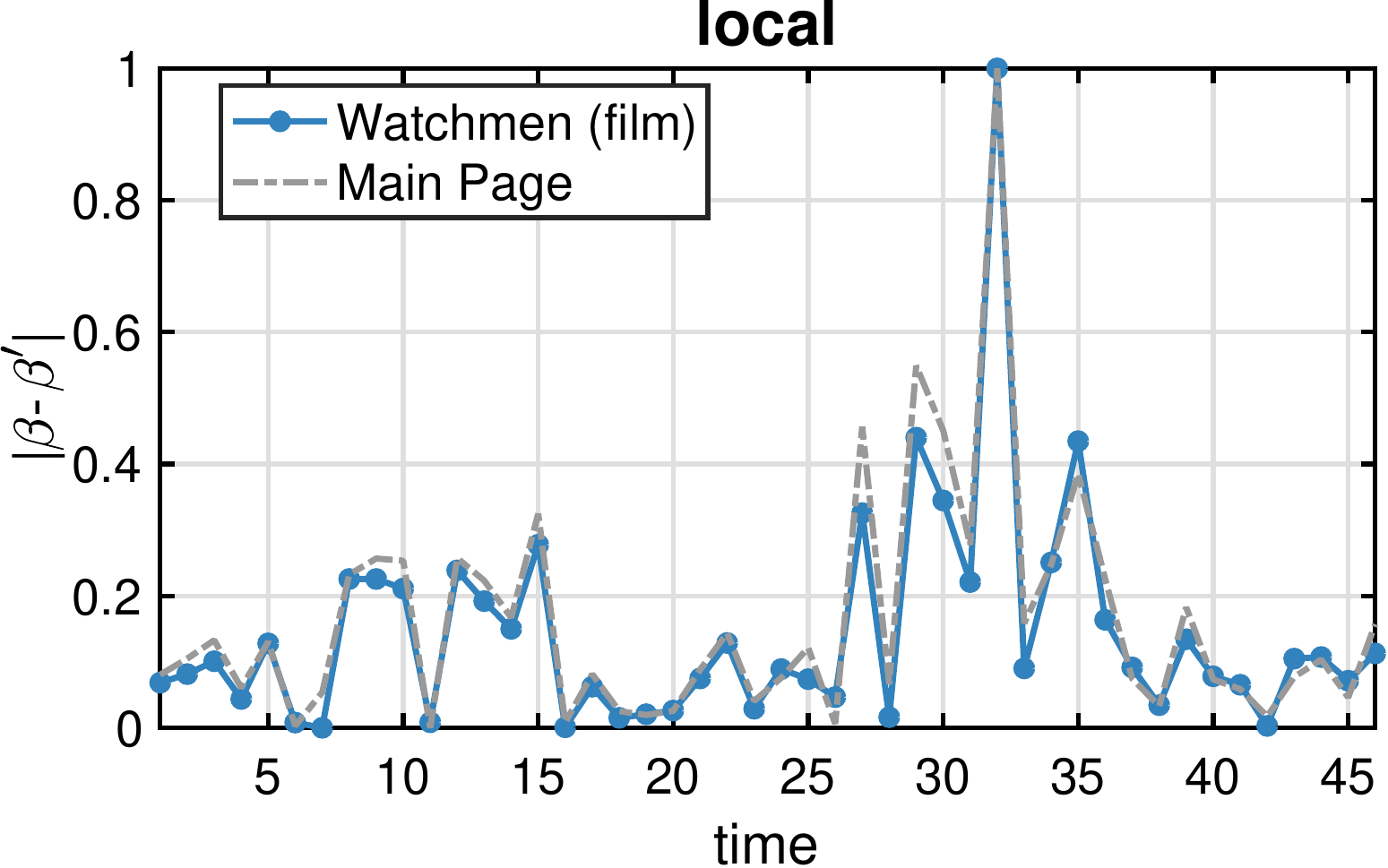}}\hfill
\subfigure{\includegraphics[width=0.245\linewidth]{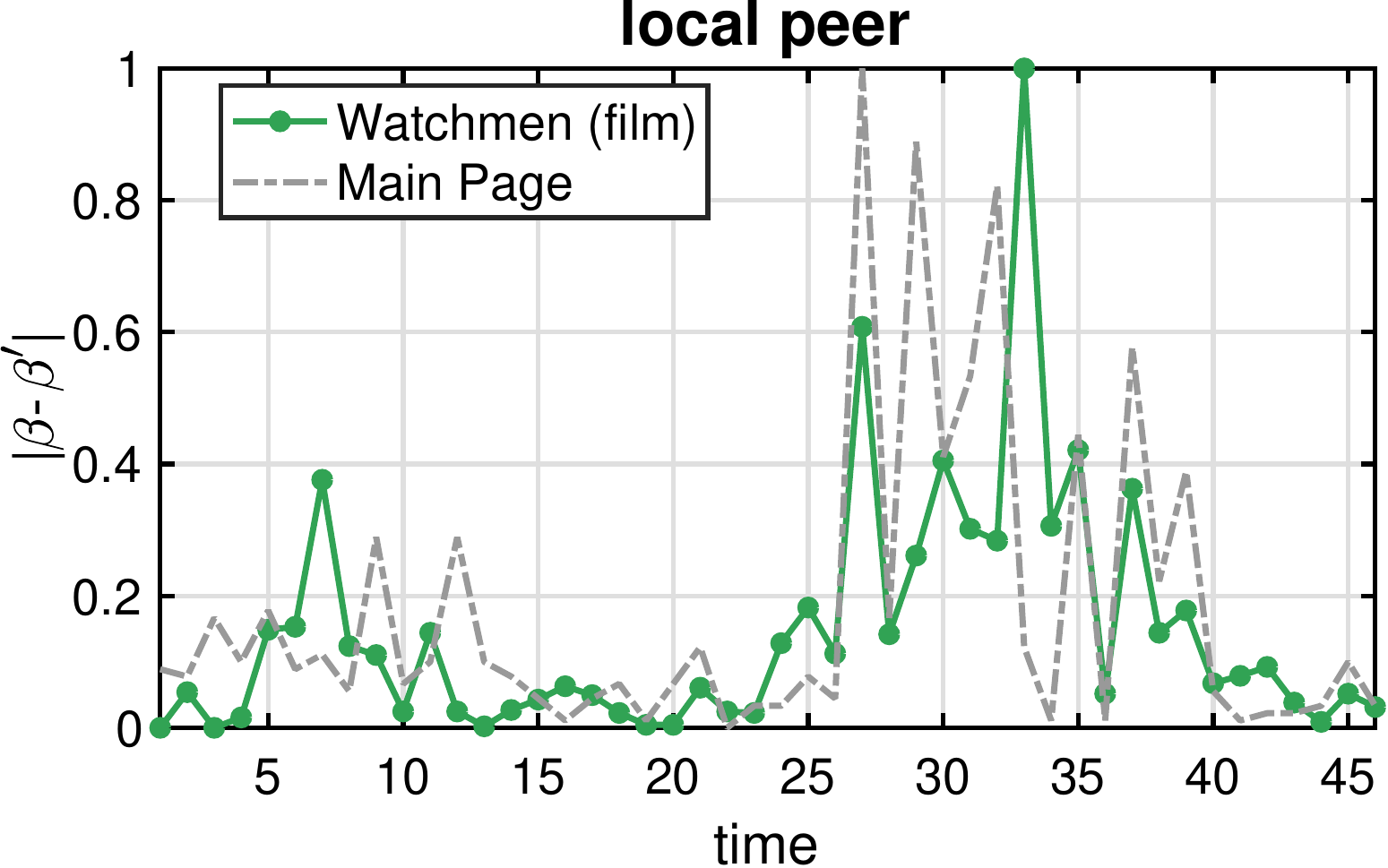}}\hfill
\subfigure{\includegraphics[width=0.245\linewidth]{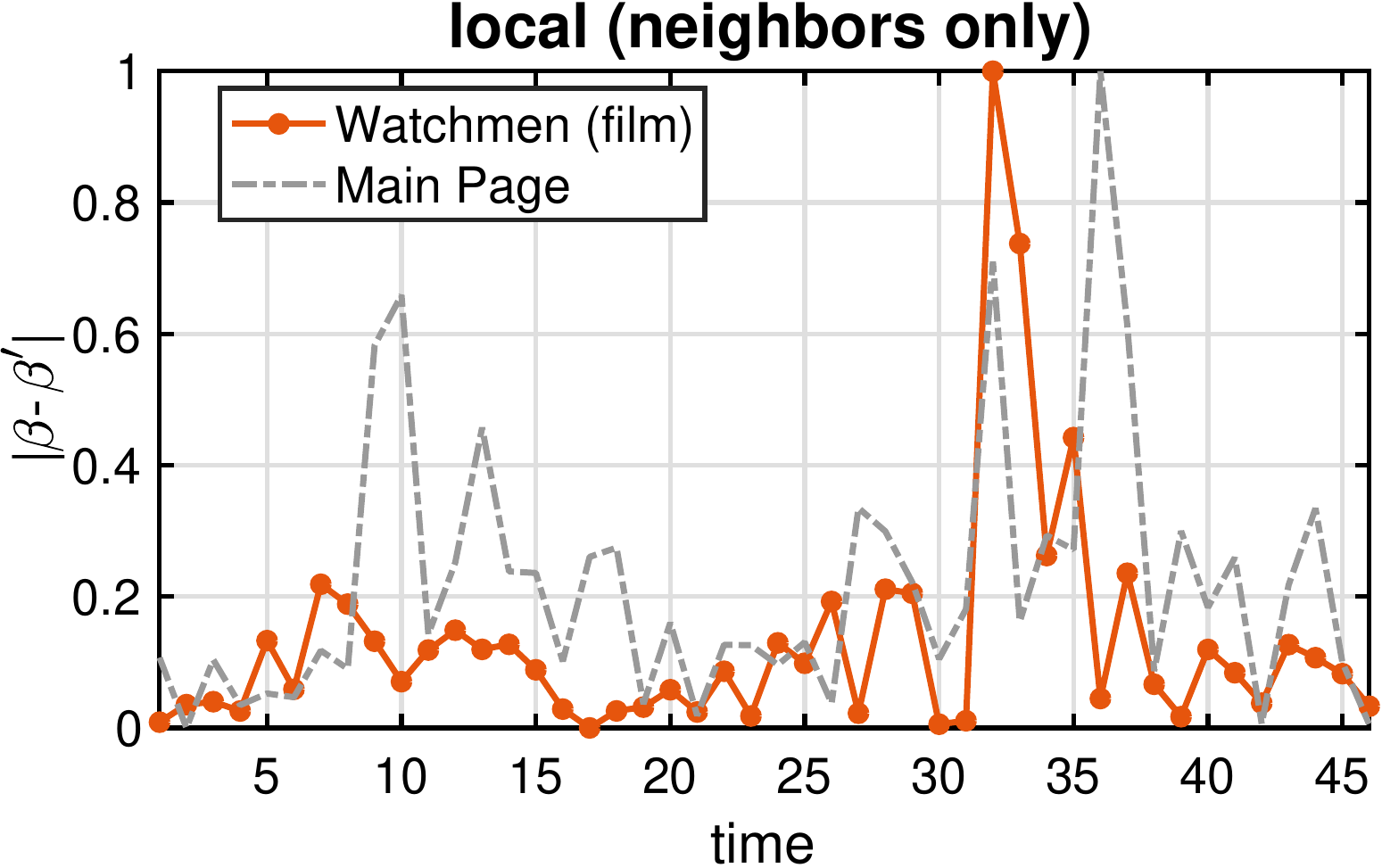}}\hfill
\subfigure{\includegraphics[width=0.245\linewidth]{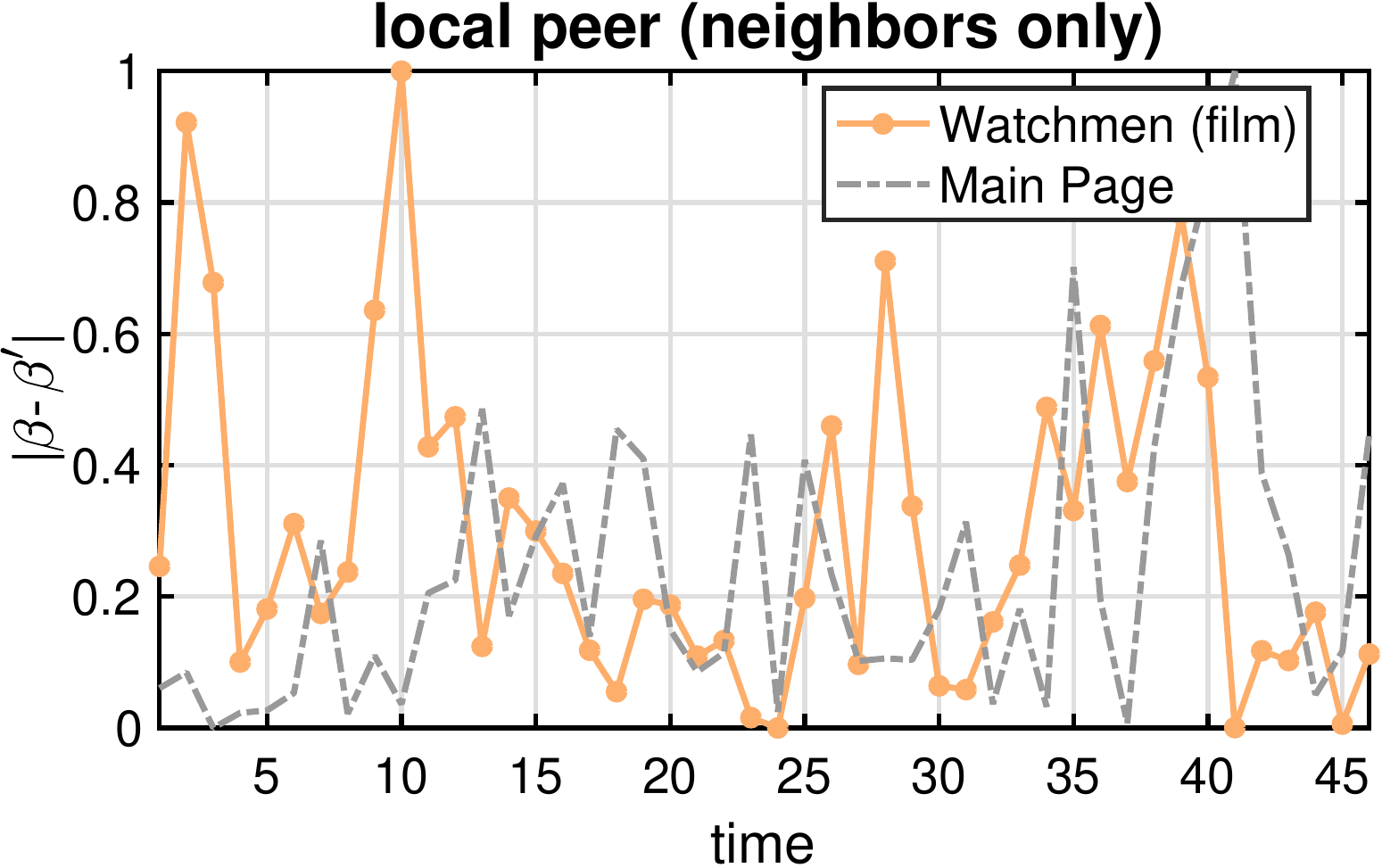}}\hfill

\caption{Comparing four different local causal inference models for a variety of Wikipedia pages, namely, the ``Main Page'', Earthquake, Seismology, Mutual Exclusion and Watchmen (film) page.
The ``Main Page'', Mutual Exclusion, and Watchmen (film) pages are unrelated to the Earthquake event that occurred off the coast of Australia. Note all time-series $\vx$ are scaled between 0 and 1 via $\nicefrac{\vx - \min(\vx)}{(\max(\vx) - \min(\vx))}$. In this experiment, we use 1 hour of data to estimate $\beta$ and the next hour to estimate $\beta^{\prime}$, and repeat this for all 48 hours by sliding the model to obtain a time-series of $\beta$'s (and $\beta^{\prime}$). See text for discussion.}
\label{fig:wiki-earthquake-time-series-of-betas}
\end{figure*}

In these experiments, we explore using the approach to infer causality between nodes that would otherwise be impossible to do without considering the graph structure and the time-series associated with each node.
One such example is shown in Figure~\ref{fig:earthquake-example-known-causal-effect}, where the effect of interest is over small timescale. 
However, with the graph structure we can hope to detect and quantify such changes in near real-time.
In particular, the page views of the Earthquake page spike at time $t$ whereas the page views of Richter Mag. appear normal at time $t$.
In the next hour, we find that the page views of Richter Magnitude spike, as shown in Figure~\ref{fig:earthquake-example-known-causal-effect}. 
This implies that immediately after the March 6, 2009 earthquake that occurred in Victoria, Australia users first visited the Earthquake page, 
and then over time, users clicked on other important and highly related pages associated with this natural disaster, such as Richter Magnitude. 
This particular case indicates the users were also interested in knowing how big the Earthquake was, a natural question after such an event.

In Figure~\ref{fig:wiki-earthquake-time-series-of-betas}, we compare four different local causal inference models for a variety of Wikipedia pages including the Earthquake, Seismology, Mutual Exclusion~(computer science), Watchmen (film), and the ``Main Page''.
The last three pages are unrelated to the Earthquake that occurred near Victoria, Australia on March 6, 2009.
Unless otherwise mentioned, we set $w=1$, $\gamma=0.1$, and a single time-step is used.
As expected, the individual causal inference model that leverages time-series from all nodes in $G$, appears very similar to the Main Page for both Earthquake and Seismology as shown in Figure~\ref{fig:wiki-earthquake-time-series-of-betas}.
Interestingly, in both cases where we estimate the causal effect using neighbors only (as opposed to the time-series from all nodes in $G$), we observe a significant difference between the time-series of $\abs{\beta - \beta^{\prime}}$ as shown in the two rightmost plots in Figure~\ref{fig:wiki-earthquake-time-series-of-betas}.

\begin{figure}[h!]
\centering
\hspace{-2mm}
\subfigure{\includegraphics[width=0.50\linewidth]{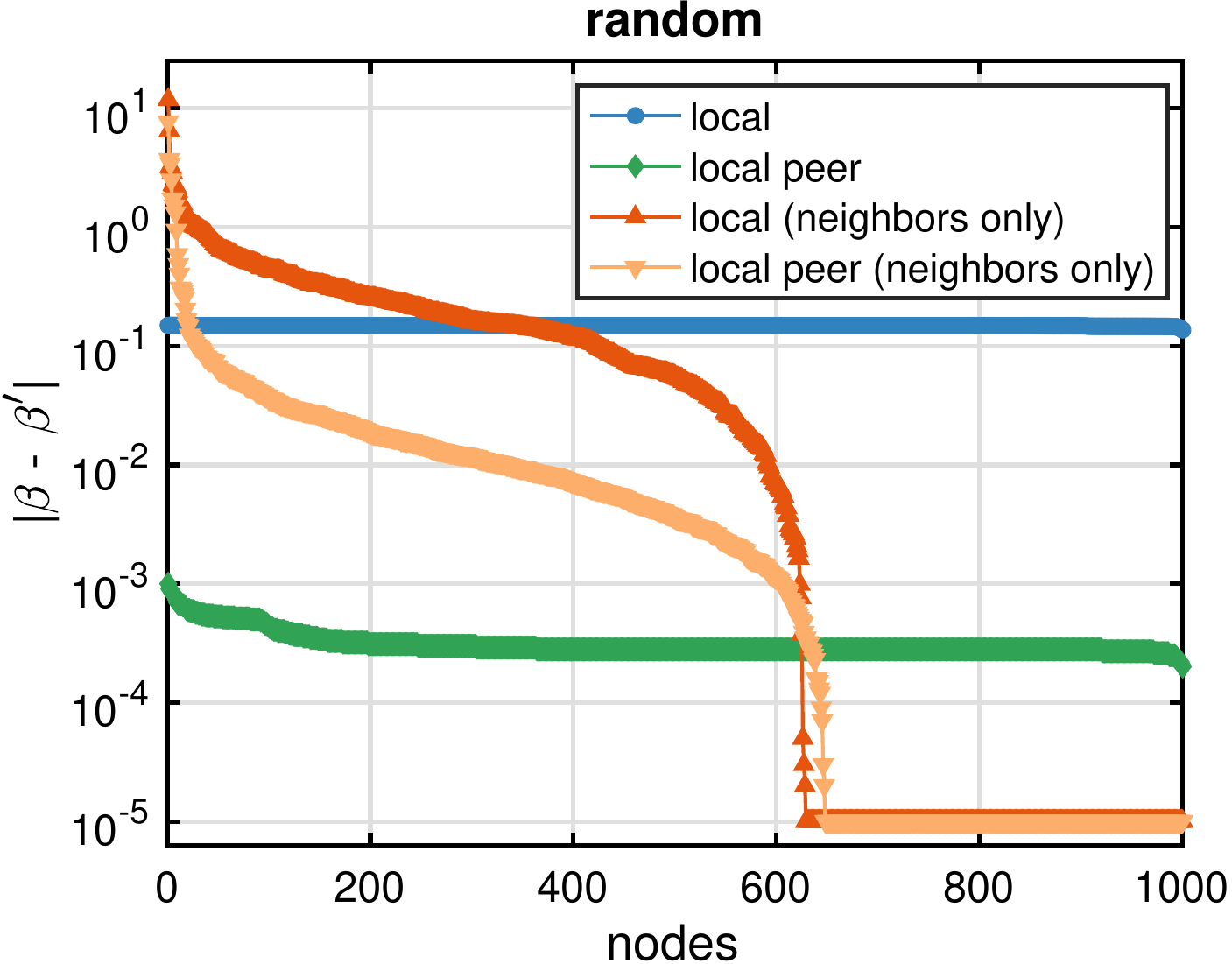}}
\subfigure{\includegraphics[width=0.50\linewidth]{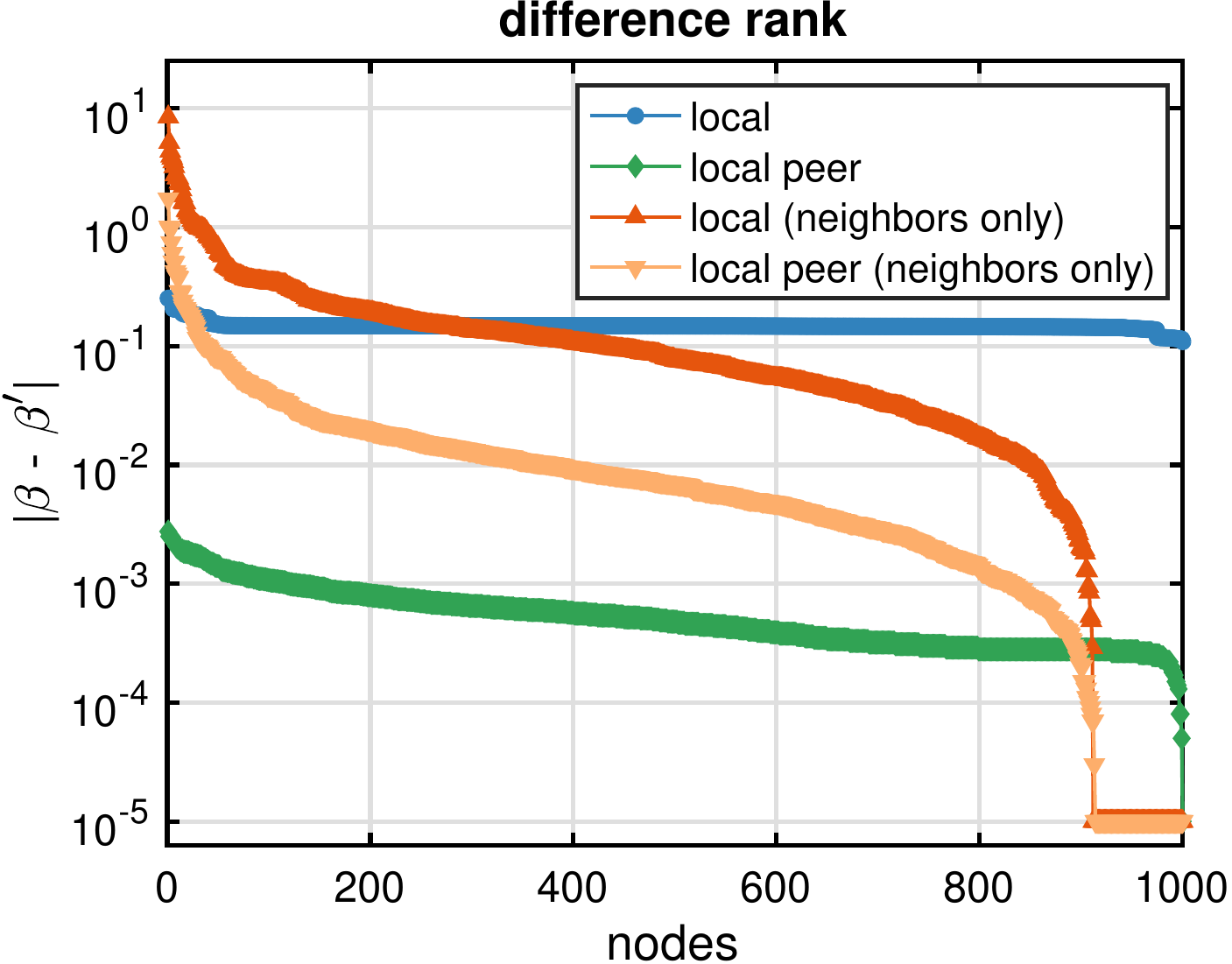}}

\vspace{-3mm}
\caption{Causal effect estimated for 1000 nodes selected uniformly at random vs. 1000 nodes with largest difference rank (Eq.~\ref{eq:diff-rank}).
See text for discussion.
}
\label{fig:diff-beta-betaprime-loglog}
\end{figure}

In the next set of experiments, we use the causal inference models to estimate the causal effects for 1000 nodes selected uniformly at random as shown in Figure~\ref{fig:diff-beta-betaprime-loglog} (top).
For comparison, we also select the top 1000 nodes given by the difference rank (Eq.~\ref{eq:diff-rank}) and use the models to estimate the causal effect for each of these individual nodes.
The difference rank of a node is the difference between its maximum and minimum value in a time-series or a restricted time window $W$ of that time-series:
\begin{equation}\label{eq:diff-rank}
    \vd = \max\limits_{t}(\vx_t) - \min\limits_{t}(\vx_t),\;\;
    \vd_W = \max\limits_{t\in W}(\vx_t) - \min\limits_{t \in W}(\vx_t)
\end{equation}\noindent
where $\vx_t \in \RR^{n}$ is the $n$-dimensional vector of page views at time $t$, \ie, $\vx_t$ is the $t$-th column of $\mX$.
The top 1000 nodes from the difference rank are the nodes with time-series that fluctuate the most.
Hence, these nodes often correspond to pages related to recent events.
This is in contrast to the majority of other Wikipedia pages with a time series that is relatively stationary with minor fluctuations.
In Figure~\ref{fig:diff-beta-betaprime-loglog}, nodes are ordered by $\abs{\beta - \beta^{\prime}}$, and for each node we estimate a $\beta$ and $\beta^{\prime}$.
As expected, $\abs{\beta - \beta^{\prime}}$ is typically smaller for nodes selected uniformly at random compared to nodes selected from the difference rank as shown in Figure~\ref{fig:diff-beta-betaprime-loglog}.
This is especially evident for the local and local peer (neighbors only) models where $\abs{\beta - \beta^{\prime}}$ is very small ($\approx\!10^{-5}$) for the last 400 nodes in Figure~\ref{fig:diff-beta-betaprime-loglog} (top; selected uniformly at random).
In contrast, only the last 100 nodes in Figure~\ref{fig:diff-beta-betaprime-loglog} (bottom) from the difference rank have $\abs{\beta - \beta^{\prime}}$ of similar magnitude.

\subsection{Effect of $\gamma$}
\begin{figure}[h!]
\centering
\subfigure{\includegraphics[width=0.60\linewidth]{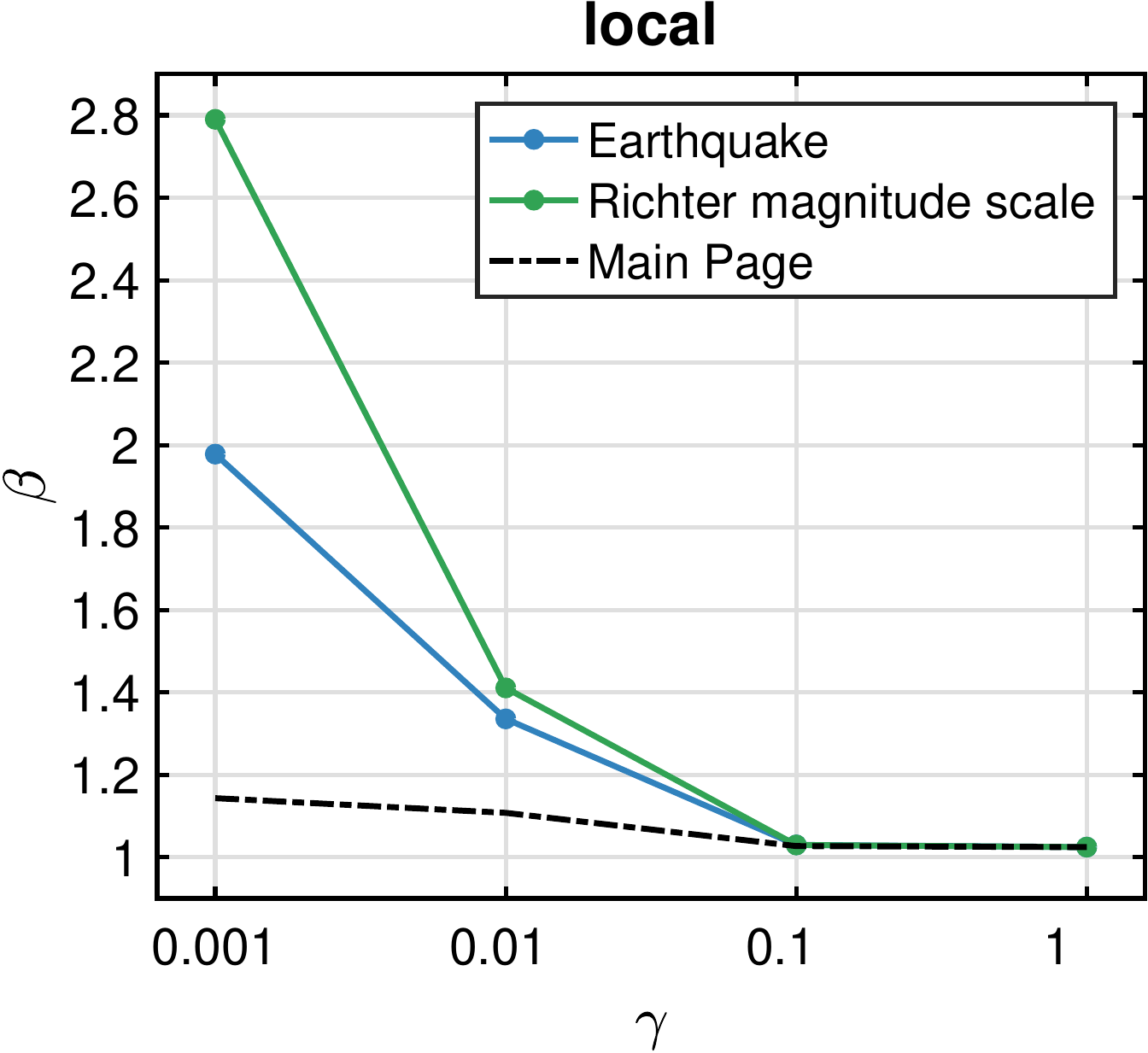}}
\subfigure{\includegraphics[width=0.60\linewidth]{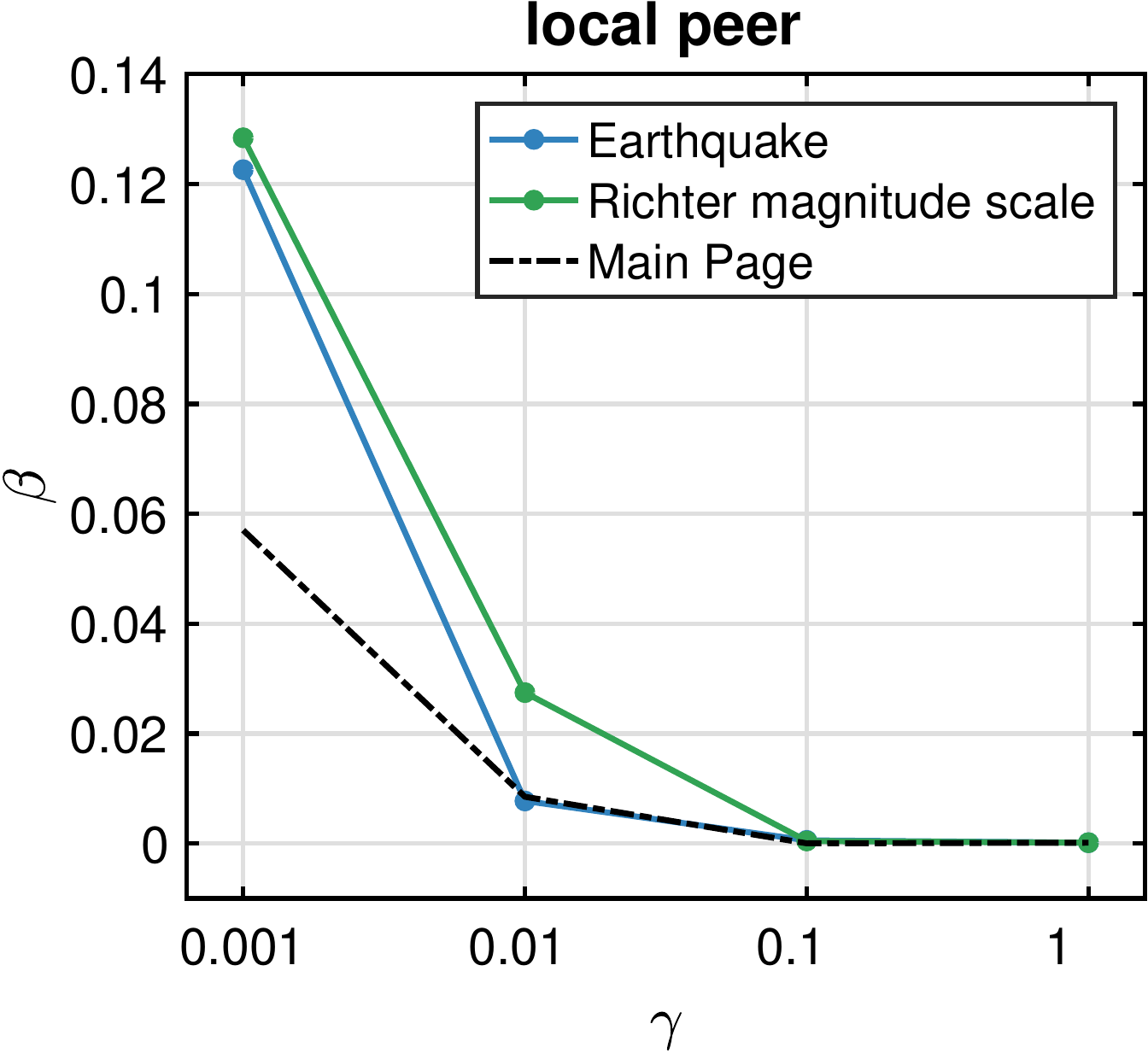}}

\vspace{-3mm}
\caption{Causal effects as $\gamma$ varies.}
\label{fig:vary-gamma}
\end{figure}

We study the effect of varying $\gamma$ for the local and local peer effect causal models.
To understand the effect of $\gamma$, we set $\gamma \in \{10^{-3}, 10^{-2}, 10^{-1}, 10^{0}\}$.
Results are shown in Figure~\ref{fig:vary-gamma}.
In particular, we study two pages with known causal effects (\ie, Earthquake and Richter magnitude scale) and the ``Main Page'' that has relatively stationary page views.
Notice that as $\gamma \rightarrow 1$, the $\beta$'s converge to the same quantity regardless of the pages and neighbors/connectivity.
Hence, as $\gamma \rightarrow 1$, the different causal models all weight the immediate neighbors the same as more distant neighbors or even nodes in different disconnected components in the graph. 
Conversely as $\gamma \rightarrow 0$, the relative difference in weight between nodes that are 1-hop away compared to nodes $k$-hops away becomes larger as shown in Figure~\ref{fig:vary-gamma}.
As an aside, when $\gamma>1$, then nodes further away in the graph from a node $i$ (larger $d(i,j)$) are given more weight than neighbors close to $i$ (\eg, immediate neighbors of $i$ are given less weight than neighbors 2-hops away and so on).

\section{Conclusion} \label{sec:conc}
Given the ubiquity of networks in modern society, developing methods for efficient inference in relational data is critical to understanding and decision making. 
The vast majority of work produced thus far ignore heterogeneity that can arise as a function of network structure. 
In this work, we studied the problem of consistent pooling observations to estimate local effects for individual nodes from relational time-series data consisting of a graph (network) where every node is associated with one or more time-series.
For this problem, we described a general approach that exploits local node-centric temporal dependencies \emph{and} topological/structural dependencies to accurately estimate effects for a given node.
We show that simpler models that do not consider the graph topology are recovered as special cases of the proposed model.
We provided a test of specification that allows practitioners to verify the consistency of the local model.
This provides practitioners the ability to verify whether the increase in power comes at the cost of a biased model.
The experiments demonstrated the effectiveness of the relational time-series causal inference models on both synthetic data with known ground-truth and a large-scale observational relational time-series data set collected from Wikipedia.

\balance
\bibliographystyle{aaai}
\bibliography{paper}

\end{document}

%% file: preamble.tex
\usepackage{comment}
\usepackage{amsmath}
\usepackage{amssymb}
\usepackage{algorithm}
\usepackage{graphicx}

\usepackage{mathtools}

\usepackage{booktabs} 
\usepackage{epstopdf}
\usepackage{comment}
\usepackage{tabularx}
\usepackage{subfigure}
\usepackage{caption}
\usepackage{tabularx}
\usepackage{paralist}
\usepackage{balance}

\usepackage{xcolor}
\definecolor{thedarkblue}{RGB}{0,0,120} 
\definecolor{mydarkblue}{rgb}{0,0.08,0.45} 
\definecolor{darkblue}{rgb}{0,0.08,180}
\colorlet{TufteRed}{red!80!black}

\definecolor{theblue}{RGB}{0,0,180}
\colorlet{thered}{TufteRed}

\usepackage{microtype}
\usepackage{balance}

\newcolumntype{L}[1]{>{\raggedright\let\newline\\\arraybackslash\hspace{0pt}}m{#1}}
\newcolumntype{C}[1]{>{\centering\let\newline\\\arraybackslash\hspace{0pt}}m{#1}}
\newcolumntype{R}[1]{>{\raggedleft\let\newline\\\arraybackslash\hspace{0pt}}m{#1}}

\usepackage{amsmath,amssymb,amsthm}

\newcommand{\eat}[1]{\ignorespaces}

\usepackage{ragged2e}
\usepackage{multirow}
\usepackage{microtype}
\usepackage{balance}
\usepackage{setspace}

\graphicspath{{./}{./graphics/}}
\newcolumntype{H}{>{\setbox0=\hbox\bgroup}c<{\egroup}@{}}

\newcolumntype{R}[1]{>{\RaggedLeft\arraybackslash}} %p{#1}}
\newcolumntype{L}[1]{>{\RaggedRight\arraybackslash}} %p{#1}}

\newcommand{\abs}[1]{\left|#1\right|}

\newcommand{\eg}{\emph{e.g.}}
\newcommand{\ie}{\emph{i.e.}}

\newtheorem{Definition}{\bfseries{Definition}}

\providecommand{\mat}[1]{\boldsymbol{\mathrm{#1}}}
\renewcommand{\vec}[1]{\boldsymbol{\mathrm{#1}}}

\DeclareMathOperator{\hugeE}{\mbox{\huge\raise-0.3ex\hbox{E}}}
\DeclareMathOperator{\p}{\mathbb{P}}
\DeclareMathOperator{\hugep}{\mbox{\huge\raise-0.3ex\hbox{$\p$}}}
\DeclareMathOperator{\Var}{Var}

\newcommand{\RR}{\mathbb{R}}

\newcommand{\ryan}[1]{\ignorespaces}

\providecommand{\mA}{\ensuremath{\mat{A}}}

\providecommand{\mD}{\ensuremath{\mat{D}}}

\providecommand{\mX}{\ensuremath{\mat{X}}}

\providecommand{\vd}{\ensuremath{\vec{d}}}
\providecommand{\ve}{\ensuremath{\vec{e}}}

\providecommand{\vx}{\ensuremath{\vec{x}}}

\graphicspath{ {./graphics/} }

\usepackage{nicefrac}